\documentclass[sn-mathphys-num, iicol]{sn-jnl}

\usepackage{tabularx}
\usepackage{booktabs}
\usepackage{array}
\usepackage[utf8]{inputenc} 
\usepackage[T1]{fontenc}    
\usepackage{url}            
\usepackage{booktabs}       
\usepackage{amsfonts}       
\usepackage{nicefrac}       
\usepackage{microtype}      
\usepackage[table,xcdraw,dvipsnames]{xcolor}         
\usepackage{graphicx}
\usepackage{amsmath}
\usepackage{pgfplots}
\pgfplotsset{compat=1.9}
\usepackage[capitalise]{cleveref}
\usepackage{pgf-pie}
\usepackage{float}
\usepackage{subcaption}
\usepackage[skip=0.333\baselineskip]{caption}
\usepackage{tikz}
\usepackage{comment}
\usepackage{amssymb}
\usepackage{colortbl}
\usepackage{epsfig}
\usepackage[inline]{enumitem}
\usepackage{soul}
\usepackage{listings}
\usepackage{algorithm}
\usepackage{numprint}
\usepackage{siunitx}
\usepackage{etoolbox}
\usepackage{cancel}
\newcommand{\ubold}{\fontseries{b}\selectfont}
\robustify\ubold
\usepackage{bbold}
\usepackage{wrapfig}
\usepackage[skip=2pt,font=scriptsize,labelfont=scriptsize]{subcaption}
\usepackage{pifont}
\usepackage{lineno}
\usepackage{array,multirow}
\usepackage[para]{footmisc}
\usepackage{pythonhighlight}

\pgfplotsset{compat=1.18}
\colorlet{punct}{red!60!black}
\definecolor{background}{HTML}{EEEEEE}
\definecolor{delim}{RGB}{20,105,176}
\colorlet{numb}{magenta!60!black}
\lstdefinelanguage{json}{
    basicstyle=\normalfont\ttfamily,
    numbers=left,
    numberstyle=\scriptsize,
    stepnumber=1,
    numbersep=8pt,
    showstringspaces=false,
    breaklines=true,
    frame=lines,
    backgroundcolor=\color{background},
    literate=
     *{0}{{{\color{numb}0}}}{1}
      {1}{{{\color{numb}1}}}{1}
      {2}{{{\color{numb}2}}}{1}
      {3}{{{\color{numb}3}}}{1}
      {4}{{{\color{numb}4}}}{1}
      {5}{{{\color{numb}5}}}{1}
      {6}{{{\color{numb}6}}}{1}
      {7}{{{\color{numb}7}}}{1}
      {8}{{{\color{numb}8}}}{1}
      {9}{{{\color{numb}9}}}{1}
      {:}{{{\color{punct}{:}}}}{1}
      {,}{{{\color{punct}{,}}}}{1}
      {\{}{{{\color{delim}{\{}}}}{1}
      {\}}{{{\color{delim}{\}}}}}{1}
      {[}{{{\color{delim}{[}}}}{1}
      {]}{{{\color{delim}{]}}}}{1},
}

\definecolor{maroon}{cmyk}{0, 0.87, 0.68, 0.32}
\definecolor{halfgray}{gray}{0.55}
\definecolor{ipython_frame}{RGB}{207, 207, 207}
\definecolor{ipython_bg}{RGB}{247, 247, 247}
\definecolor{ipython_red}{RGB}{186, 33, 33}
\definecolor{ipython_green}{RGB}{0, 128, 0}
\definecolor{ipython_cyan}{RGB}{64, 128, 128}
\definecolor{ipython_purple}{RGB}{170, 34, 255}

\lstdefinelanguage{iPython}{
    morekeywords={access,and,break,class,continue,def,del,elif,else,except,exec,finally,for,from,global,if,import,in,is,lambda,not,or,pass,print,raise,return,try,while},%
    morekeywords=[2]{abs,all,any,basestring,bin,bool,bytearray,callable,chr,classmethod,cmp,compile,complex,delattr,dict,dir,divmod,enumerate,eval,execfile,file,filter,float,format,frozenset,getattr,globals,hasattr,hash,help,hex,id,input,int,isinstance,issubclass,iter,len,list,locals,long,map,max,memoryview,min,next,object,oct,open,ord,pow,property,range,raw_input,reduce,reload,repr,reversed,round,set,setattr,slice,sorted,staticmethod,str,sum,super,tuple,type,unichr,unicode,vars,xrange,zip,apply,buffer,coerce,intern},%
    sensitive=true,%
    morecomment=[l]\#,%
    morestring=[b]',%
    morestring=[b]",%
    morestring=[s]{'''}{'''},
    morestring=[s]{"""}{"""},
    morestring=[s]{r'}{'},
    morestring=[s]{r"}{"},%
    morestring=[s]{r'''}{'''},%
    morestring=[s]{r"""}{"""},%
    morestring=[s]{u'}{'},
    morestring=[s]{u"}{"},%
    morestring=[s]{u'''}{'''},%
    morestring=[s]{u"""}{"""}%
    literate=
    {á}{{\'a}}1 {é}{{\'e}}1 {í}{{\'i}}1 {ó}{{\'o}}1 {ú}{{\'u}}1
    {Á}{{\'A}}1 {É}{{\'E}}1 {Í}{{\'I}}1 {Ó}{{\'O}}1 {Ú}{{\'U}}1
    {à}{{\`a}}1 {è}{{\`e}}1 {ì}{{\`i}}1 {ò}{{\`o}}1 {ù}{{\`u}}1
    {À}{{\`A}}1 {È}{{\'E}}1 {Ì}{{\`I}}1 {Ò}{{\`O}}1 {Ù}{{\`U}}1
    {ä}{{\"a}}1 {ë}{{\"e}}1 {ï}{{\"i}}1 {ö}{{\"o}}1 {ü}{{\"u}}1
    {Ä}{{\"A}}1 {Ë}{{\"E}}1 {Ï}{{\"I}}1 {Ö}{{\"O}}1 {Ü}{{\"U}}1
    {â}{{\^a}}1 {ê}{{\^e}}1 {î}{{\^i}}1 {ô}{{\^o}}1 {û}{{\^u}}1
    {Â}{{\^A}}1 {Ê}{{\^E}}1 {Î}{{\^I}}1 {Ô}{{\^O}}1 {Û}{{\^U}}1
    {œ}{{\oe}}1 {Œ}{{\OE}}1 {æ}{{\ae}}1 {Æ}{{\AE}}1 {ß}{{\ss}}1
    {ç}{{\c c}}1 {Ç}{{\c C}}1 {ø}{{\o}}1 {å}{{\r a}}1 {Å}{{\r A}}1
    {€}{{\EUR}}1 {£}{{\pounds}}1
    {^}{{{\color{ipython_purple}\^{}}}}1
    {=}{{{\color{ipython_purple}=}}}1
    {+}{{{\color{ipython_purple}+}}}1
    *{-}{{{\color{ipython_purple}-}}}1
    {*}{{{\color{ipython_purple}$^\ast$}}}1
    {/}{{{\color{ipython_purple}/}}}1
    {+=}{{{+=}}}1
    {-=}{{{-=}}}1
    {*=}{{{$^\ast$=}}}1
    {/=}{{{/=}}}1,
    identifierstyle=\color{black}\ttfamily,
    commentstyle=\color{ipython_cyan}\ttfamily,
    stringstyle=\color{ipython_red}\ttfamily,
    keepspaces=true,
    showspaces=false,
    showstringspaces=false,
    rulecolor=\color{ipython_frame},
    frame=single,
    frameround={t}{t}{t}{t},
    framexleftmargin=6mm,
    numbers=left,
    numberstyle=\tiny\color{halfgray},
    backgroundcolor=\color{ipython_bg},
    basicstyle=\scriptsize,
    keywordstyle=\color{ipython_green}\ttfamily,
}

\begin{document}

\title[Article Title]{Public Health Advocacy Dataset: A Dataset of Tobacco Usage Videos from Social Media}


\author*[1]{\fnm{Naga VS Raviteja} \sur{Chappa}}\email{nchappa@uark.edu}

\author[2]{\fnm{Charlotte} \sur{McCormick}}\email{cem044@uark.edu}

\author[2]{\fnm{Susana Rodriguez} \sur{Gongora}}\email{sr069@uark.edu}

\author[2]{\fnm{Page Daniel} \sur{Dobbs}}\email{pdobbs@uark.edu}


\author[1]{\fnm{Khoa} \sur{Luu}}\email{khoaluu@uark.edu}

\affil*[1]{\orgdiv{Department of EECS}, \orgname{University of Arkansas}, \orgaddress{ \city{Fayetteville} \state{AR}, \country{USA}}}

\affil[2]{\orgdiv{Center for Public Health and Technology}, \orgname{University of Arkansas}, \orgaddress{ \city{Fayetteville} \state{AR}, \country{USA}}}
\website{\emph{Project Website}: \url{https://uark-cviu.github.io/projects/PHAD/}}



\abstract{
The Public Health Advocacy Dataset (PHAD) is a comprehensive collection of 5,730 videos related to tobacco products sourced from social media platforms like TikTok and YouTube. This dataset encompasses 4.3 million frames and includes detailed metadata such as user engagement metrics, video descriptions, and search keywords. This is the first dataset with these features providing a valuable resource for analyzing tobacco-related content and its impact. Our research employs a two-stage classification approach, incorporating a Vision-Language (VL) Encoder, demonstrating superior performance in accurately categorizing various types of tobacco products and usage scenarios. The analysis reveals significant user engagement trends, particularly with vaping and e-cigarette content, highlighting areas for targeted public health interventions. The PHAD addresses the need for multi-modal data in public health research, offering insights that can inform regulatory policies and public health strategies. This dataset is a crucial step towards understanding and mitigating the impact of tobacco usage, ensuring that public health efforts are more inclusive and effective.}


\maketitle

\section{Introduction}

In the rapidly evolving landscape of digital content, short-form videos on platforms like TikTok and YouTube have revolutionized user-generated content, reshaping how information is consumed and shared. This shift has created an urgent need for advanced deep learning applications capable of comprehending the nuances of video content, particularly in addressing emerging public health concerns.

While Large Language Models (LLMs) have shown remarkable capabilities in processing textual data, their adaptation to video understanding remains limited. Current state-of-the-art LLM architectures~\cite{liu2024visual, zhu2023minigpt, li2022blip, li2023blip, dai2024instructblip} excel in natural language processing but lack explicit support for video inputs. This limitation is particularly pronounced on platforms where video content is the primary mode of communication and expression.

The vast and diverse user bases of platforms like YouTube and TikTok have transformed them into microcosms of content ranging from entertainment to social commentary. However, this digital ecosystem has inadvertently facilitated the promotion of tobacco-related products, raising significant public health concerns~\cite{lee2023cigarette, kwon2020perceptions, jancey2023promotion, sun2023vaping}. Existing LLMs, designed primarily for text-based operations, are ill-equipped to address this issue effectively, as they neglect the rich visual information embedded in video content.

Our research addresses this critical gap by proposing an innovative approach at the intersection of deep learning and video analysis. Focusing on the nuanced domain of tobacco-related content, we present a two-stage end-to-end system that extracts essential cues from videos and trains a vision-language model to comprehend and analyze the contextual intricacies within video content. By combining the power of deep learning with the unique challenges presented by video-centric formats, our approach aims to provide a comprehensive solution to the complex issue of tobacco promotion in this dynamic digital space.

\noindent Our key contributions are as follows:

\begin{itemize}
    \item \textbf{Public Health Advocacy Dataset (PHAD)}: We introduce a novel, comprehensive dataset of tobacco-related videos from YouTube and TikTok. This dataset includes rich metadata such as user engagement metrics, video descriptions, and search keywords, providing a robust foundation for research in public health and content moderation.

    \item \textbf{Balanced Sampling Strategy}: We employ an innovative approach to dataset construction, using TikTok videos as negative samples (containing tobacco product usage or promotion) and YouTube videos as positive samples (without tobacco product usage or promotion). This balanced strategy enhances the study of tobacco usage, addiction, and promotion across different social media platforms.

    \item \textbf{Two-Stage Classification Framework}: We propose a novel classification method that integrates a Vision-Language (VL) Encoder. This two-stage approach significantly improves the accuracy of classifying tobacco-related content by leveraging visual and textual features, addressing the limitations of current text-centric language models in video analysis.

    \item \textbf{User Engagement Analysis}: Our research uncovers significant trends in user interaction with tobacco-related content, particularly highlighting the high engagement with vaping and e-cigarette videos. These insights provide crucial information for targeted public health interventions and content moderation strategies.

    \item \textbf{Contextual Feature Evaluation}: We demonstrate the critical role of contextual features, such as metadata, in enhancing the performance of deep learning models for video content understanding and classification. This contribution underscores the importance of a holistic video analysis approach beyond visual content alone.
\end{itemize}

Through these contributions, our work addresses the pressing need for advanced deep learning applications in video content analysis, particularly in the context of public health concerns related to tobacco promotion on social media platforms. By combining novel dataset creation, innovative modeling techniques, and comprehensive analysis, we provide a foundation for future research and practical applications in content moderation and public health advocacy.

This paper unfolds our proposed methodology and expected results, shedding light on the imperative role that advanced deep learning applications can play in addressing emerging challenges in video content moderation and public health initiatives.

\vspace{-3mm}
\section{Related Work}

We compare our dataset with the prior published datasets, shown in \cref{tab:dataset_comparison}.

\begin{table*}[ht]
    \centering
    \caption{Comparison of PHAD dataset with prior related works.}
    \label{tab:dataset_comparison}
    \small 
    \renewcommand{\arraystretch}{1.0} 
    \begin{tabularx}{\textwidth}{@{}l@{\hskip 3pt}c@{\hskip 3pt}c@{\hskip 3pt}c@{\hskip 3pt}c@{\hskip 3pt}c@{\hskip 3pt}c@{}}
        \toprule
        \textbf{Dataset Name} & \textbf{Videos} & \textbf{Frames} & \textbf{\begin{tabular}[c]{@{}c@{}}Social Media\\ Platforms\end{tabular}} & \textbf{Classes} & \textbf{\begin{tabular}[c]{@{}c@{}}Features\end{tabular}} & \textbf{License} \\ 
        \midrule
        Murthy et. al~\cite{ntad184} & 254 & 826 & TikTok & 3 & \begin{tabular}[c]{@{}c@{}}Single tobacco \\ product (Vape)\end{tabular} & Academic research \\
        Vassey et. al~\cite{ntad224} & 3,837 & 6,999 & \begin{tabular}[c]{@{}c@{}}TikTok and \\ Instagram\end{tabular} & 7 & \begin{tabular}[c]{@{}c@{}}Multiple tobacco \\ products\end{tabular} & Academic research \\ 
        \rowcolor{green} 
        \textbf{PHAD (Ours)} & \textbf{5,730} & \textbf{4.3M} & \textbf{\begin{tabular}[c]{@{}c@{}}TikTok and \\ YouTube\end{tabular}} & \textbf{8} & \textbf{\begin{tabular}[c]{@{}c@{}}User Engagement Metrics, \\ Tobacco Product Types, \\ and Metadata features\end{tabular}} & \textbf{\begin{tabular}[c]{@{}c@{}} CC BY-NC-SA 4.0, \\ Academic research\end{tabular}} \\ 
        \bottomrule
    \end{tabularx}
\end{table*}

Recent advancements in video analysis~\cite{wang2021actionclip, chappa2023sogar, chappa2023spartan, tang2018vctree, zhong2024learning, zareian2020bridging, zareian2020learning, lu2016visual, zhong2021learning, ye2021linguistic, nguyen2024type, chappa2024flaash, truong2022otadapt, jalata2022eqadap, chappa2020squeeze, chappa2024advanced, chappa2024public} have predominantly employed traditional deep learning algorithms, but the absence of language comprehension limits their efficacy. Recognizing this gap, there is a shift towards integrating text modalities~\cite{chappa2024react, chappa2024hatt} to enhance understanding capabilities. It underscores the need to evolve existing algorithms, advocating for joint training with Large Language Models (LLMs)~\cite{liu2024visual} to achieve superior performance and nuanced comprehension. Our work builds upon this paradigm shift, proposing a novel approach seamlessly integrating video and language understanding. Through leveraging the power of LLMs, our framework bridges the gap between visual and textual cues, advancing the state-of-the-art in video analysis. This convergence enables a holistic interpretation of video content, opening new avenues for more accurate, context-aware applications.

Recently~\cite{kong2023understanding, murthy2023influence, chappa2024advanced} used machine learning techniques to understand, specifically on YouTube, how the user profile attributes influence e-cigarette searches and analysis of their content and promotion strategies. Prior studies~\cite{ntad224} on e-cigarette content in social media have relied on qualitative and text-based machine learning methods. They developed an image-based computer vision model using a unique dataset of 6,999 Instagram images labeled for eight object classes related to e-cigarettes, demonstrating an innovative approach to automatically and scalably monitor tobacco promotions online. Murthy \emph{et al.}~\cite{ntad184} have relied on text-based analyses to detect e-cigarette content on social media, which addressed the visual nature of platforms like TikTok by developing a computer vision model that uses YOLOv7~\cite{wang2023yolov7}. This model was trained and tested on a dataset of 826 images, annotated for vaping devices, hands, and vapor clouds, extracted from 254 TikTok posts identified via relevant hashtags, showcasing a practical approach for monitoring e-cigarette content on visual social media platforms.

\section{PHAD Description}


The dataset annotation pipeline and data format are detailed in~\cref{annotation_pipeline} and~\cref{data_format}. This section outlines the methodology for sourcing and curating video data into a meaningful machine learning dataset to enhance understanding of tobacco product usage and addiction analysis across diverse demographics. The "PHAD" is briefly described below, including its contents and the ethical considerations involved in its collection and use.

\begin{figure*}[ht]
    \centering
    \begin{minipage}{0.3\textwidth}
        \centering
        \includegraphics[width=0.5\textwidth]{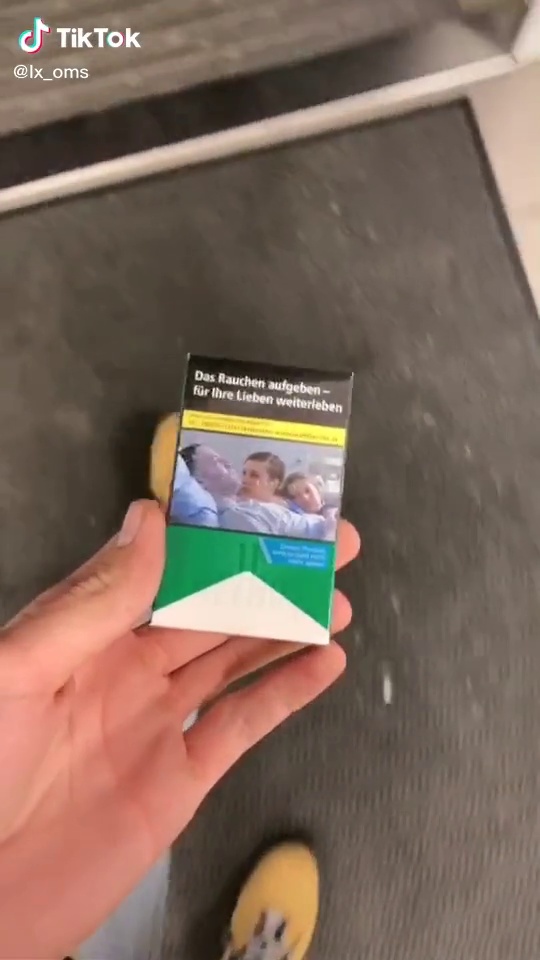}
        \caption*{(a) Cigarettes}
    \end{minipage}
    \begin{minipage}{0.3\textwidth}
        \centering
        \includegraphics[width=0.5\textwidth]{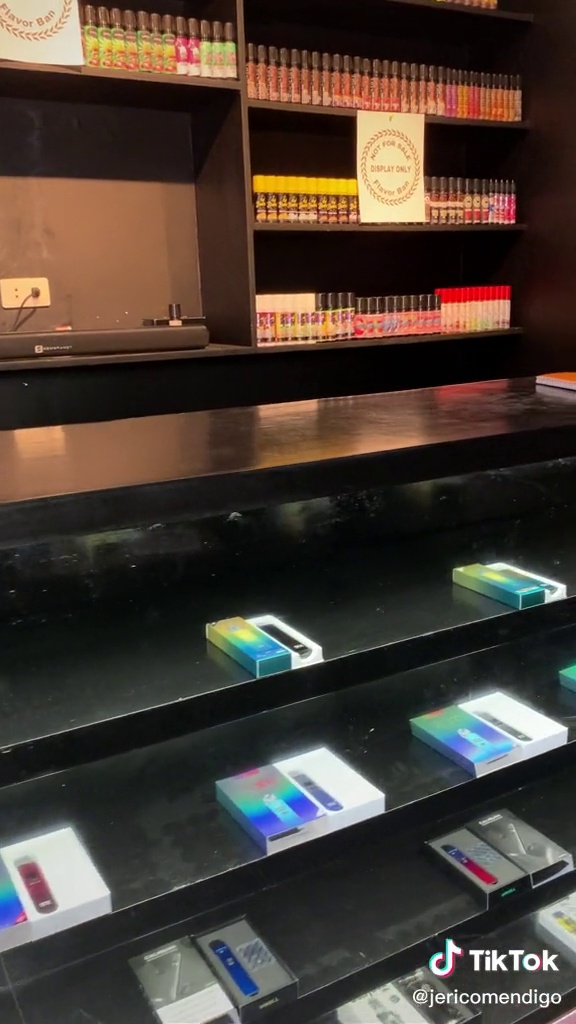}
        \caption*{(b) E-cigarettes}
    \end{minipage}
    \begin{minipage}{0.3\textwidth}
        \centering
        \includegraphics[width=0.5\textwidth]{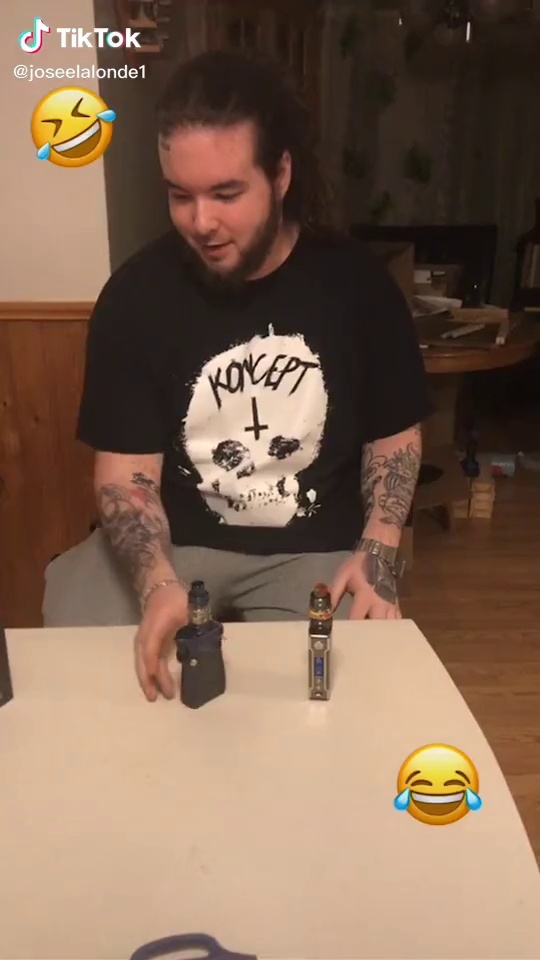}
        \caption*{(c) Vaping Devices}
    \end{minipage}
    \begin{minipage}{0.3\textwidth}
        \centering
        \includegraphics[width=0.5\textwidth]{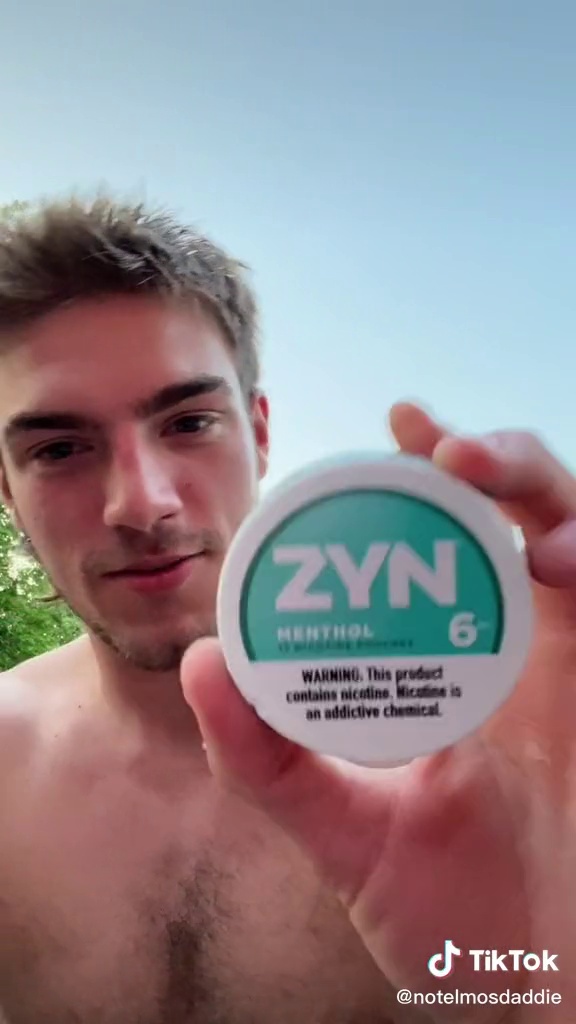}
        \caption*{(d) Smokeless Tobacco}
    \end{minipage}
    \begin{minipage}{0.3\textwidth}
        \centering
        \includegraphics[width=0.5\textwidth]{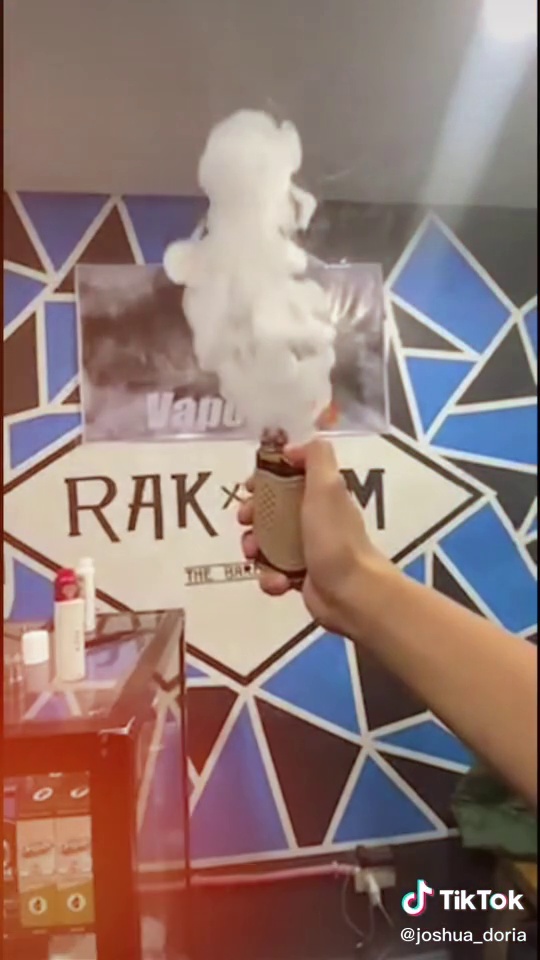}
        \caption*{(e) Vape with Smoke}
    \end{minipage}
    \begin{minipage}{0.3\textwidth}
        \centering
        \includegraphics[width=0.5\textwidth]{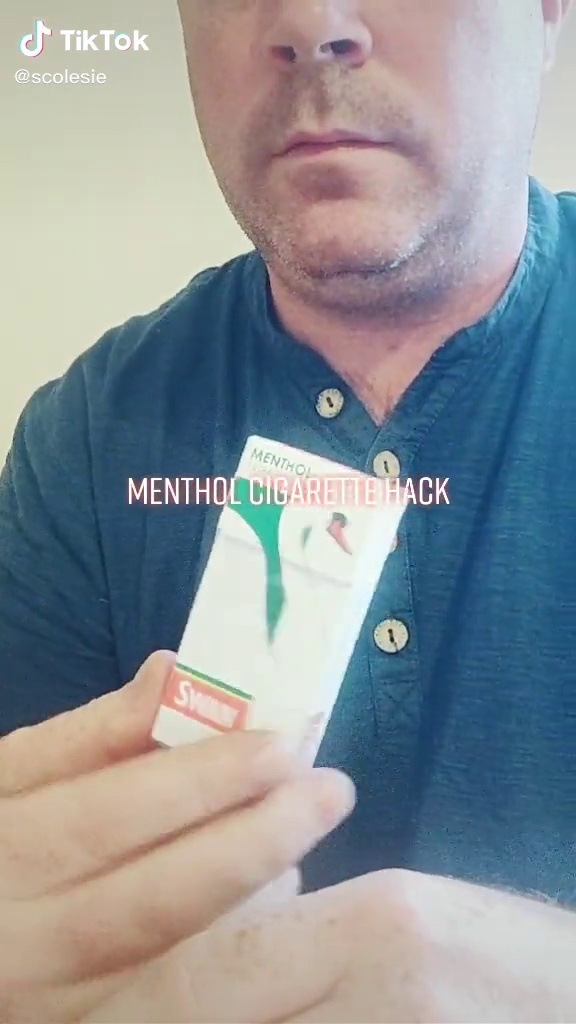}
        \caption*{(f) Menthol Cigarettes}
    \end{minipage}
    \caption{Sample images from the dataset categorized by tobacco product type.}
    \label{fig:dataset_samples}
\end{figure*}

\subsection{Contents}
This dataset comprises a collection of 5,730 videos, with approximately 4,014 sourced from YouTube~\cite{yt} and 1,716 from TikTok~\cite{tiktok}. These platforms were chosen due to their popularity and the vast amount of user-generated content related to tobacco products. The videos in the dataset represent a wide array of content types, including promotional material, personal vlogs, educational content, and public health advertisements. The videos are annotated with several features aimed at facilitating a multifaceted analysis of tobacco use:
\begin{itemize}

    \item \textbf{User Engagement Metrics}: Contains the likes, shares, comments, and views of the videos in the dataset.
    
    \item \textbf{Type of Tobacco Products}: Classifies the tobacco product shown or discussed in the video (e.g., cigarettes, vaping devices, chewing tobacco).
    
    \item \textbf{Search Keywords}: The keywords used to locate the video may indicate common search behaviors and public interest in specific tobacco products.


    \item \textbf{Video Descriptions}: Includes a summary of the video content, providing context and additional details not readily apparent from video analysis alone.

\end{itemize}

The dataset totals over \textbf{12.3 GB} (videos only) in size and is stored in widely used formats suitable for video processing. The metadata for each video includes a unique identifier, the original video URL, and tagged features, as outlined above.
\subsection{Licensing}
The videos collected fall under various licenses, with many available under standard content licenses provided by YouTube~\cite{yt} and TikTok~\cite{tiktok}. For this dataset, we focus on videos that allow for academic and non-commercial use to ensure compliance with licensing agreements. The use of these videos for educational and research purposes aligns with the terms of service of the hosting platforms, which typically permit such use.

Our dataset is licensed under the CC BY-NC-SA 4.0 license, "Creative Commons Attribution-NonCommercial-ShareAlike 3.0." This license allows users to share and adapt the dataset, provided that they give appropriate credit, do not use the material for commercial purposes, and distribute their contributions under the same license.

The CC BY-NC-SA 4.0 license is well-suited for our dataset, as it is intended solely for research and non-commercial purposes. This license includes works licensed under CC-BY (Creative Commons Attribution), CC-BY-SA (Creative Commons Attribution-ShareAlike), and CC-BY-NC (Creative Commons Attribution-NonCommercial). These licenses collectively ensure that: (1) Users can share the dataset. (2) Users can adapt the dataset with appropriate credit. By adhering to these licensing terms, we ensure that our machine learning dataset can be used and adapted freely within the research community while respecting the original creators' conditions.

\subsection{Data Source}
The videos were identified using a combination of automated scraping techniques~\cite{api} and manual curation to ensure a broad representation of content related to tobacco products. This approach allowed for the collection of diverse videos, reflecting different contexts and cultural perspectives of tobacco usage and its implications.

This dataset was curated to provide a resource for researchers in public health, psychology, and machine learning. It aims to facilitate a deeper understanding of how tobacco products are portrayed on social media and the impact of these portrayals on public perception and behavior. 
\subsection{Ethical Considerations}
This dataset consists of publicly available videos collected from YouTube and TikTok. Given the public nature of these platforms, individual content creators' consent for each video was not explicitly obtained; however, our data collection and usage strictly adhere to the platforms' terms of service regarding public content.

We have implemented rigorous data handling procedures to ensure the dataset is ethically used, particularly for educational and research purposes. It includes:-
\textbf{Compliance with Platform Policies}: All data collection has been performed in compliance with the terms of service of the respective platforms~\cite{yt,tiktok}, which allow the analysis of publicly shared content for academic and research purposes. \textbf{Use Limitation}: The dataset is intended solely for non-commercial use, mainly focused on academic research and public health studies. Researchers are encouraged to apply further ethical considerations in their use of the data, especially when publishing results that could impact individuals or communities.

Furthermore, the dataset excludes any content that might involve sensitive personal data or content that could compromise the privacy and dignity of individuals without their explicit consent. This careful curation process ensures that the dataset is both a valuable academic resource and a respectful tool that acknowledges the ethical implications of using publicly sourced video content.

\section{Dataset Characterization}
    
This section examines the videos in our dataset to provide insights into their raw characteristics, focusing on user engagement, tobacco product types, and associated metadata. This diversity of content types and metadata across platforms helps address potential biases in public health research models.

\subsection{Videos by User Engagement}
Analyzing user engagement metrics reveals how frequently users interact with tobacco-related content on YouTube and TikTok. These metrics, including views, likes, comments, and shares, indicate the content's reach and impact, reflecting public interest and influence.
\textbf{Views:} The number of views highlights the content's visibility and reach. Higher views suggest a broader influence on the audience regarding tobacco use.
\textbf{Likes and Dislikes:} These provide a measure of viewer sentiment, offering insights into public approval or disapproval of the depicted tobacco products or usage.
\textbf{Comments:} Analyzing comments reveals viewer opinions, misconceptions, and discussions, critical for understanding public perceptions and social discourse around tobacco use.
\textbf{Shares:} The frequency of shares indicates the video's capacity to engage viewers and encourage them to propagate its content, suggesting higher engagement levels and potential viral spread.

\begin{figure*}
    \centering
    \includegraphics[width=0.8\textwidth]{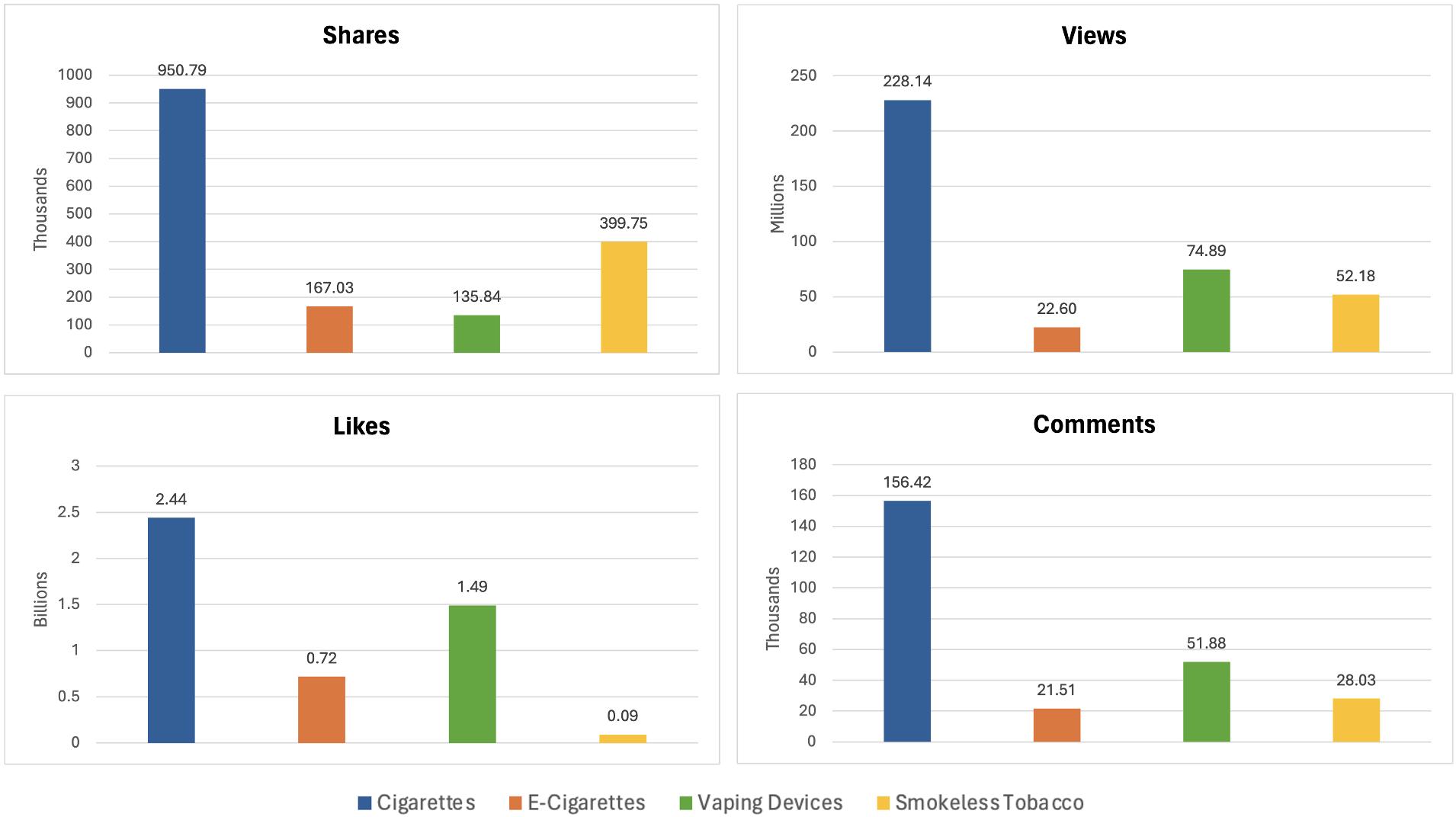}
    \caption{Engagement metrics categorized by the type of tobacco product discussed. \textbf{Best viewed in zoom and color.}}
    \label{fig:engagement_metrics}
\end{figure*}

\cref{fig:engagement_metrics} could illustrate the average number of views, likes, comments, and shares per video, categorized by the type of tobacco product discussed. This visualization would allow for a comparative analysis of how tobacco products stimulate viewer interaction and engagement.

\subsection{Videos by Tobacco Product Type}
Videos are categorized based on the type of tobacco product featured or discussed. The categorization includes cigarettes, e-cigarettes, vaping devices, and smokeless tobacco products, among others. This classification helps understand specific trends and the influence of different product types on consumer behavior. \cref{fig:tobacco_product_distribution} illustrates the percentage distribution of videos by tobacco product type, highlighting a predominant focus on {vaping devices and e-cigarettes }, indicative of trending consumption patterns among young adults, teens, and adults demographics.

\subsection{Videos by Metadata Features}
Each video in the dataset is associated with metadata that includes search keywords, price range, and video descriptions. These metadata features are crucial for contextual analysis and are leveraged to enhance the interpretability of deep learning models to study public health implications.

\textbf{Search Keywords}: Keywords provide insights into standard terms and phrases individuals use when searching for tobacco-related content. This data helps understand public interest and awareness levels about different tobacco products.
\textbf{Video Descriptions}: Descriptions add a layer of qualitative data that can be analyzed to extract thematic elements and sentiment, providing deeper insights into the narrative around tobacco use portrayed in the videos.

\begin{figure}[ht]
    \centering
    \begin{minipage}{0.45\textwidth}
        \centering
        \includegraphics[width=0.95\textwidth]{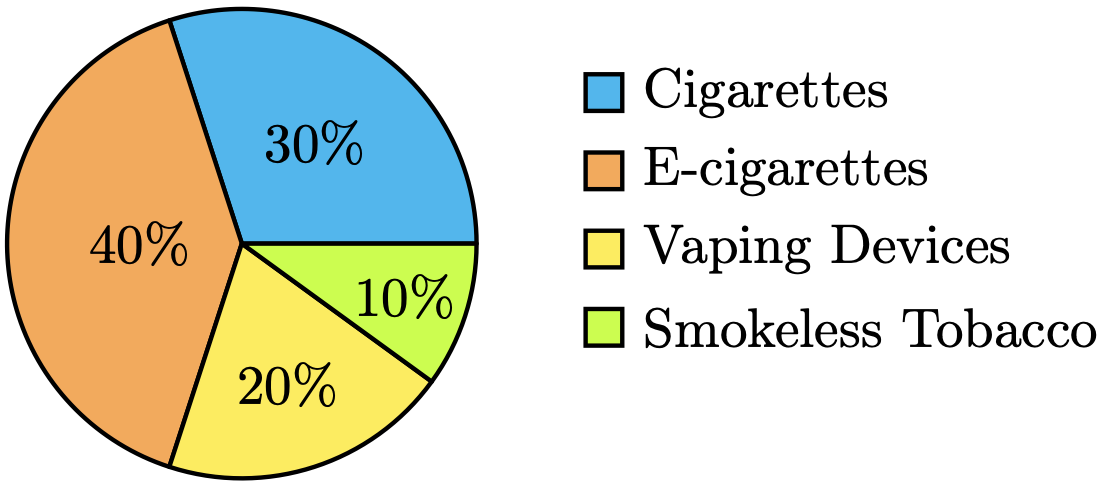}
        \caption{Percentage distribution of videos by tobacco product type.}
        \label{fig:tobacco_product_distribution}
    \end{minipage}
    \hfill
    \begin{minipage}{0.5\textwidth}
        \centering
        \includegraphics[width=0.85\textwidth]{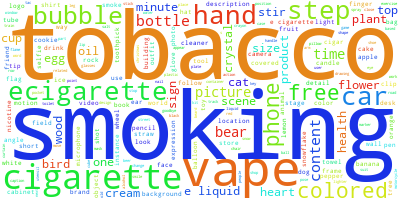}
        \caption{Different search keywords are presented as a word cloud. }
        \label{fig:videos_per_keyword_category}
    \end{minipage}
\end{figure}

\cref{fig:videos_per_keyword_category} displays a word cloud of the search keywords used to scrape the data, emphasizing the most frequently used search terms associated with the videos in the dataset.

\subsection{Video Characterization}
In addition to the qualitative metadata, quantitative video data analysis, such as length and resolution, is also conducted. Most of the videos are in HD resolution, reflecting the modern standard of video content on social media platforms. The diversity in video length, ranging from quick 30-second TikTok videos to longer YouTube tutorials or reviews, indicates how information is consumed and shared across different demographics.

\cref{tab:video_characteristics} provides a breakdown of video characteristics, showing the minimum, maximum, and median values for video length and resolution across the dataset. This table helps understand the dataset's capacity to provide comprehensive visual information suitable for deep learning models.

\begin{table}[ht]
    \centering
    \caption{Video data characteristics}
    \label{tab:video_characteristics}
    \normalsize
    \begin{tabular}{@{}l@{\hspace{4pt}}c@{\hspace{4pt}}c@{\hspace{4pt}}c@{}}
        \toprule
        \textbf{Characteristic} & \textbf{Min} & \textbf{Max} & \textbf{Avg} \\
        \midrule
        Length (s) & 3 & 1,500 & 120 \\
        Resolution & 360p & 1080p & 720p \\
        \bottomrule
    \end{tabular}
\end{table}

\textbf{Note:} We present a clear and comprehensive \emph{Dataset use cases} section in~\cref{dataset_use_cases}.



\section{Dataset Limitations}\label{sec:limitations}
This section discusses the inherent limitations associated with our dataset due to technical constraints and dataset-specific characteristics. Understanding these limitations is vital for interpreting the findings derived from this dataset accurately and for guiding future enhancements.

\subsection{Technical Limitations}
\textbf{Video Quality Variability:} Despite most videos being in HD resolution, there is significant variability in video quality due to factors like compression, lighting, and camera quality. Such variability can affect the performance of video analysis algorithms, particularly those involving {action recognition and object detection}.

\textbf{Data Incompleteness:} Not all videos include comprehensive metadata such as accurate price ranges or complete descriptions. This incompleteness can lead to gaps in analyses, particularly in understanding the full context of tobacco use depicted in the videos.

\textbf{Platform-Specific Biases}: The videos sourced from YouTube and TikTok reflect the content guidelines and user demographics of these platforms, which may not be representative of broader or different social media ecosystems. It can limit the generalizability of the findings to other platforms or offline behaviors.

\subsection{Dataset-Specific Limitations}
\textbf{Under Representation of Certain Demographics:} While the dataset includes videos from a wide range of sources, certain demographic groups, especially those from lower socioeconomic backgrounds or less represented regions, might be underrepresented. It can skew the analysis towards the behaviors and preferences of more active internet users or those from specific geographical areas.

\textbf{Content Creator Bias:} Videos created for public viewing are often curated or produced with specific messages in mind, which can introduce bias regarding how tobacco products are portrayed. It might affect the dataset's utility in objectively analyzing tobacco product usage and public perceptions.

\section{Evaluation}\label{sec:methods}

This section introduces our two-stage pipeline for video content classification, utilizing Resnet-50~\cite{he2016deep} as the backbone network for feature extraction.

\subsection{Stage One: Feature Extraction}
In the first stage, discriminative features are extracted from input video frames. Video frames \( f_t, f_{t+1}, \ldots, f_{t+N} \) are processed independently through the Resnet-50 network to produce feature maps \( x_t, x_{t+1}, \ldots, x_{t+N} \), capturing essential spatial characteristics for subsequent classification.

The feature extraction is mathematically represented as:
\begin{equation}
 x_t = F(f_t; \theta)
\end{equation}
where \( F \) is the feature extractor, parameterized by \( \theta \).

\subsection{Stage Two: Feature Classification}
The second stage involves classifying the extracted features using various temporal modeling techniques. We explore five different approaches:

\noindent\textbf{Vanilla Approach:}
Features \( x_t \) are directly fed into a simple classifier without temporal modeling.
\begin{equation}
 y_t = C(x_t; \phi)
\end{equation}

\noindent\textbf{Handcrafted Window:}
Temporal information is integrated by applying a predefined kernel \( K \) over the features.
\begin{equation}
 \begin{array}{ll}
 \hat{x}_t = \sum_{i=-m}^{m} K(i) \cdot x_{t+i} & \\
 y_t = C(\hat{x}_t; \phi)
 \end{array}
\end{equation}

\noindent\textbf{1D Convolutional Layer:}
A 1D CNN captures temporal dependencies by sliding a convolutional filter over the feature set.
\begin{equation}
 \begin{array}{ll}
 \hat{x}_t = \sum_{i=-m}^{m} w_i \cdot x_{t+i} & \\
 y_t = C(\hat{x}_t; \phi)
 \end{array}
\end{equation}

\noindent\textbf{Transformer Encoder~\cite{vaswani2017attention}:}
A Transformer Encoder models complex dependencies between time steps using self-attention.
\begin{equation}
 \begin{array}{ll}
 \hat{x}_t = \text{Attention}(Q(x_t), K(X), V(X)) & \\
 y_t = C(\hat{x}_t; \phi)
 \end{array}
\end{equation}

Her, Q, K, and V are queries, keys, and values of the corresponding features.
\noindent\textbf{Vision-Language Encoder~\cite{jia2021scaling}:}
This encoder provides a contextually enriched classification by combining visual features \( x_t \) with textual descriptors \( z_t \).
\begin{equation}
 \begin{array}{ll}
 \hat{x}_t = H(x_t, z_t; \psi) & \\
 y_t = C(\hat{x}_t; \phi)
 \end{array}
\end{equation}

Here, \( C \) is the classifier, \( H \) is the fusion function, and \( \phi \) and \( \psi \) are learned parameters during training.

\subsection{Experimental Results}
The experimental setup and evaluation metrics used are detailed in~\cref{experimental_setup} and~\cref{evaluation_metrics}. The following sections present the experimental results and ablation study analysis. We also performed additional experiments which are presented in~\cref{additional_experiments}.

\subsubsection{Benchmarking and Comparisons with Existing Datasets}

To provide a more robust comparison with existing datasets, we conducted an additional experiment comparing PHAD to the dataset from Murthy et al.~\cite{ntad184}. To facilitate a fair comparison with their three-class dataset, we adapted PHAD by consolidating our eight classes into three broader categories: (1) Vaping devices, (2) Traditional tobacco products, and (3) Non-tobacco content. We then compared the performance of our model on this adapted version of PHAD with the reported results from Murthy et al. \cite{ntad184}.

Table \ref{tab:comparison} presents the performance metrics for the Murthy et al.~\cite{ntad184} dataset, our adapted 3-class PHAD, and the full 8-class PHAD.

\begin{table}[h]
\centering
\caption{Performance comparison across datasets.}
\label{tab:comparison}
\begin{tabular}{@{}l@{\hspace{3pt}}c@{\hspace{3pt}}c@{\hspace{3pt}}c@{\hspace{3pt}}c@{}}
\hline
\textbf{Dataset} & \textbf{Acc.} & \textbf{Prec.} & \textbf{Rec.} & \textbf{F1} \\
\hline
Murthy et al. \cite{ntad184} & - & 86.3\% & 77.1\% & 81.4\% \\
Adapted PHAD (3 classes) & 89.5\% & 90.8\% & 91.1\% & 89.4\% \\
Full PHAD (8 classes) & 73.2\% & 65.2\% & 67.5\% & 71.8\% \\
\hline
\end{tabular}
\raggedright
    \footnotesize
    Note: \textbf{Acc.}: Accuracy, \textbf{Prec.}: Precision, \textbf{Rec.}: Recall, \textbf{F1}: F1-score. 
\end{table}

The results demonstrate that our model performs exceptionally well on the 3-class adaptation, surpassing the reported results for the Murthy et al.~\cite{ntad184} dataset. This performance highlights the quality of our data and annotations. As expected, performance decreases with the full 8-class classification, reflecting the increased complexity and granularity of the task. However, this more nuanced classification offers greater utility for detailed tobacco content analysis.

This experiment provides a comprehensive view of PHAD's capabilities in relation to existing datasets. It demonstrates both our dataset's scalability and its potential for more granular analysis when used in its complete form. The ability to adapt PHAD for different classification granularities while maintaining high performance underscores its value to the research community, offering flexibility for various analytical needs in tobacco-related content studies.

\begin{table}[ht]
    \centering
    \caption{Performance comparison of methods on the test set of PHAD.}
    \label{tab:performance_comparison}
    \normalsize 
    \begin{tabular}{@{}l@{\hspace{3pt}}c@{\hspace{3pt}}c@{\hspace{3pt}}c@{\hspace{3pt}}c@{}}
        \toprule
        \textbf{Method} & \textbf{Acc.} & \textbf{Prec.} & \textbf{Rec.} & \textbf{F1} \\
        \midrule
        CNN+LSTM~\cite{hochreiter1997long} & 54.2 & 51.6 & 52.3 & 47.8 \\
        I3D~\cite{carreira2017quo} & 61.7 & 56.2 & 54.1 & 59.8 \\
        \textbf{Ours} & \textbf{73.2} & \textbf{65.2} & \textbf{67.5} & \textbf{71.8} \\
        \bottomrule
    \end{tabular}
    \raggedright
    \footnotesize
    Note: \textbf{Acc.}: Accuracy, \textbf{Prec.}: Precision, \textbf{Rec.}: Recall, \textbf{F1}: F1-score. All values are percentages.
\end{table}

\subsubsection{Comparison with Baseline Methods}

We compared our proposed two-stage method with the following baseline methods:
\textbf{ConvNet + LSTM~\cite{hochreiter1997long}}: A supervised model trained on manually labeled data.
\textbf{I3D~\cite{carreira2017quo}}: A transfer learning model using pre-trained weights from a similar domain.
More information about the baseline methods is provided in Supplementary material. 
Table \ref{tab:performance_comparison} presents the performance of our two-stage method compared to the baseline methods on the test set. Our two-stage method outperformed the baseline methods across all evaluation metrics. The higher F1 score indicates a better balance between precision and recall.

\subsubsection{Comparison of Different Approaches in Stage-2 of Our Framework}

In the second stage of our pipeline, we evaluated five approaches to capturing temporal dynamics and context, with the Vision-Language (VL) Encoder outperforming the others.

\textbf{Vanilla Approach:} This baseline model, which directly classifies extracted features without temporal modeling, provided moderate performance, serving as a reference point.
\textbf{Handcrafted Window:} By applying a predefined kernel to integrate temporal information, this approach improved performance over the vanilla model by emphasizing significant frames.
\textbf{1D Convolutional Layer:} Utilizing a 1D CNN to learn temporal dependencies, this method further enhanced performance, effectively capturing temporal relationships within the feature set.
\textbf{Transformer Encoder:} Leveraging self-attention mechanisms, the Transformer Encoder excelled at modeling complex dependencies across time steps, showing significant improvement in classification accuracy, especially for longer sequences.
\textbf{Vision-Language Encoder:} Combining visual features with textual descriptors, the VL Encoder achieved the best performance. This approach provided a comprehensive understanding of video content by processing annotated text labels alongside visual cues, resulting in the most contextually enriched classification outputs.

\begin{figure}[ht]
    \centering
    \includegraphics[width=0.5\textwidth]{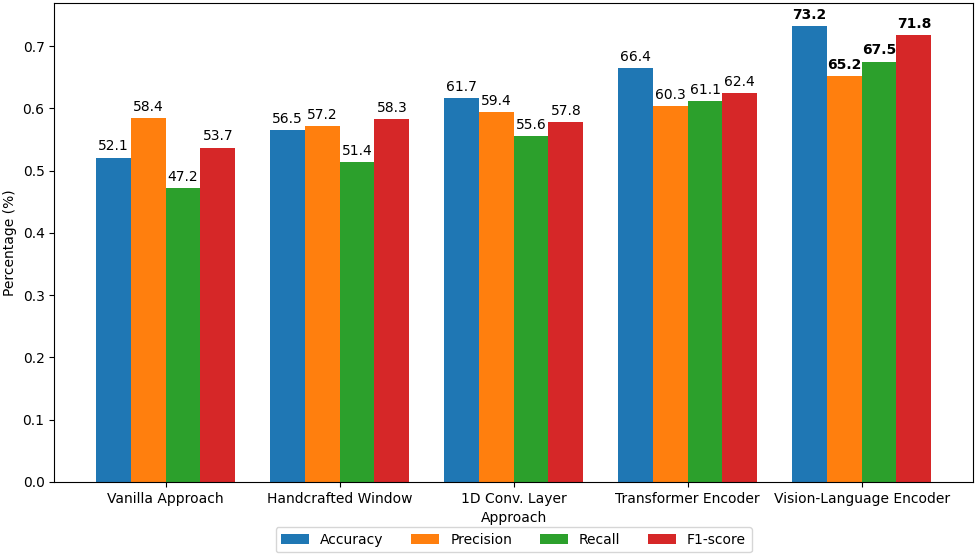}
    \caption{Comparison of different approaches in the second stage of our framework. \textbf{Best viewed in color and zoom.}}
    \label{fig:stage_two_comparison}
\end{figure}

\cref{fig:stage_two_comparison} summarizes the performance of different approaches used in the second stage of our framework. 

\subsection{Ablation Study}

\textbf{Significance of 2-Stage Approach:}
To understand the impact of the two-stage process, we conducted an ablation study comparing the single-stage and two-stage approaches. \cref{tab:ablation_study} shows the performance of the single-stage feature extraction alone versus the two-stage approach that includes the Vision-Language Encoder.


\begin{table}[ht]
    \centering
    \small 
    \begin{minipage}{0.47\textwidth}
        \centering
        \caption{Ablation study results comparing single-stage and two-stage approaches. Both experiments are performed with ResNet-50~\cite{he2016deep} as the backbone.}
        \label{tab:ablation_study}
        \begin{tabular}{@{}l@{\hskip 3pt}c@{\hskip 3pt}c@{\hskip 3pt}c@{\hskip 3pt}c@{}}
            \toprule
            \textbf{Approach} & \textbf{Accuracy} & \textbf{Precision} & \textbf{Recall} & \textbf{F1 score} \\
            \midrule
            Single-Stage & 61.7\% & 56.2\% & 54.1\% & 59.8\% \\
            Two-Stage &\textbf{73.2\%} & \textbf{65.2\%} & \textbf{67.5\%} & \textbf{71.8\%}  \\
            \bottomrule
        \end{tabular}
    \end{minipage}
    \hspace{0.01\textwidth}
    \begin{minipage}{0.47\textwidth}
        \centering
        \caption{Ablation study results comparing the impact of adding each proposed feature.}
        \label{tab:feature_ablation_study}
        \begin{tabular}{@{}l@{\hskip 3pt}c@{\hskip 3pt}c@{\hskip 3pt}c@{\hskip 3pt}c@{}}
            \toprule
            \textbf{Features} & \textbf{Accuracy} & \textbf{Precision} & \textbf{Recall} & \textbf{F1 score} \\
            \midrule
            Base Only & 51.2\% & 49.7\% & 50.1\% & 48.6\% \\
            + Temporal & 61.7\% & 56.2\% & 54.1\% & 59.8\% \\
            + Visual & 64.3\% & 59.3\% & 57.1\% & 62.4\% \\
            + Textual & \textbf{73.2\%} & \textbf{65.2\%} & \textbf{67.5\%} & \textbf{71.8\%}  \\
            \bottomrule
        \end{tabular}
    \end{minipage}
\end{table}
The results indicate that the two-stage approach, which includes the Vision-Language (VL) Encoder, significantly outperforms the single-stage feature extraction approach. It highlights the critical role of the VL Encoder in enhancing the model's ability to capture and utilize contextual information for improved classification performance.

\textbf{Significance of Proposed Features:}
To further demonstrate the importance of the proposed features, we conducted another ablation study to evaluate the performance improvement by adding each feature incrementally. ~\cref{tab:feature_ablation_study} shows how adding each feature impacts the overall performance.


The incremental addition of features shows a consistent improvement in performance metrics:
\textbf{Base Features Only}: Basic features extracted in the first stage without any enhancements. 
\textbf{+ Temporal Features}: Adding temporal features to capture sequence dependencies.
\textbf{+ Visual Context}: Including additional visual context to enhance spatial relationship understanding.
\textbf{+ Textual Descriptions}: Incorporating textual descriptions for comprehensive context, leading to the best performance with the complete set of proposed features.

\section{Discussion}

Our analysis of the PHAD reveals significant insights into tobacco-related content on social media. This section discusses our results, dataset limitations, and future research directions.





\noindent\textbf{Implications for Public Health and Policy:}
Insights from PHAD have several implications for public health and policy:-

\emph{Targeted Interventions}: Focus on vaping and e-cigarette content for educational campaigns and regulatory measures to curb use among younger populations.

\emph{Enhanced Monitoring and Regulation}: Accurate classification aids regulators in monitoring compliance with advertising laws and identifying illegal tobacco sales or underage use.

\emph{Informing Policy Decisions}: Policymakers can use these insights to inform tobacco control strategies and public health campaigns.

\noindent\textbf{Future Work:}
Future research should address limitations and explore additional avenues:-

\emph{Expanding the Dataset}: Include data from more social media platforms and regions for a comprehensive view.

\emph{Improving Data Quality}: Enhance metadata quality and completeness through sophisticated data collection and annotation methods.

\emph{Advanced Analysis Techniques}: Explore advanced machine learning techniques, such as reinforcement learning and generative models, to improve content classification.

\emph{Longitudinal Studies}: Conduct studies to understand trends in tobacco use and public perception over time.

\section{Conclusions}

We introduced the Public Health Advocacy Dataset, a comprehensive dataset containing videos related to tobacco products sourced from YouTube and TikTok. This dataset included detailed metadata such as user engagement metrics, video descriptions, and search keywords, providing a rich resource for analyzing tobacco-related content and its impact.
Our analysis demonstrated that the PHAD significantly improved the performance of classification tasks, especially when using a two-stage method incorporating a Vision-Language (VL) Encoder. This approach effectively leveraged visual and textual features, enhancing the model's ability to categorize various tobacco products and usage scenarios accurately.
This dataset provided valuable insights into public engagement trends, such as the high interaction with vaping and e-cigarette content, which is crucial for informing public health interventions and regulatory policies. The PHAD contributed to a more nuanced understanding of tobacco use and its portrayal on social media by addressing the need for multi-modal data in public health research.
To that end, the PHAD has been a significant step forward in providing researchers and policymakers with the tools needed to better understand and address the challenges posed by tobacco use. The dataset's diverse and comprehensive nature helped combat biases and ensure public health strategies are more inclusive and effective.

\backmatter

\bmhead{Supplementary information} Our dataset is publicly availble on \url{https://uark-cviu.github.io/projects/PHAD/}. Please contact the corresponding author if there is any issue accessing the website/downloading the dataset.

\bmhead{Acknowledgements}
We would like to acknowledge Arkansas Biosciences Institute (ABI) Grant, and NSF Data Science, Data Analytics that are Robust and Trusted (DART) for their funding in supporting this research.

\section*{Declarations}

\noindent \textbf{Funding:} This work is supported by NSF DART Award \#1946391.

\noindent \textbf{Conflict of interest/Competing interests:} Not applicable

\noindent \textbf{Ethics approval and consent to participate:} Not applicable

\noindent \textbf{Consent for publication:} Not applicable

\noindent \textbf{Data availability:} Please refer to \textit{Supplementary information} for more details on data availability.  

\noindent \textbf{Materials availability:} Not applicable

\noindent \textbf{Code availability:} Implementation of the proposed framework will be released upon the acceptance of the article. 

\noindent \textbf{Author contribution:} Conceptualization, N.V.S.R.C., P.D.D. and K.L.; Data Curation, N.V.S.R.C, C.M. and S.R.G.; Data pre-processing, N.V.S.R.C.; methodology, N.V.S.R.C.; programming, N.V.S.R.C.; validation, N.V.S.R.C., and K.L.; formal analysis, N.V.S.R.C.; investigation, K.L.; resources, K.L.; writing---original draft preparation, N.V.S.R.C., C.M. and S.R.G.; writing---review and editing, P.D.D., and K.L.; visualization, N.V.S.R.C.; supervision, P.D.D., and K.L.. All authors have read and agreed to the published version of the manuscript.






\begin{appendices}

\section{The PHAD Dataset Datasheet}

The original questions are in \textbf{bold}. The subtext to each question is in italics. The answers are in plain text with no formatting. The questions were copied from the 'Datasheets for Datasets' paper available online: \url{https://arxiv.org/pdf/1803.09010.pdf}.

\noindent\text{\large{Motivation}}

\emph{The questions in this section are primarily intended to encourage dataset creators to clearly articulate their reasons for creating the dataset and to promote transparency about funding interests.}

\noindent\textbf{For what purpose was the dataset created?}

\noindent\emph{Was there a specific task in mind? Was there a specific gap that needed to be filled? Please provide a description.}

The PHAD was created to address the growing need for comprehensive data on tobacco-related content across social media platforms, specifically YouTube and TikTok. The dataset was developed with the following objectives in mind:

1. Provide a Rich Resource for Analyzing Tobacco-Related Content:
   PHAD includes diverse videos related to various tobacco products, accompanied by detailed metadata such as user engagement metrics, video descriptions, and search keywords. This comprehensive dataset allows researchers to analyze the nature and impact of tobacco-related content on social media.

2. Enhance Public Health Research:
   By offering a multi-modal dataset that includes both visual and textual features, PHAD aims to improve the accuracy and effectiveness of models used in public health research. The dataset helps in understanding user engagement trends and the influence of tobacco-related content, which is crucial for developing targeted public health interventions.

3. Fill the Gap in Existing Datasets:
   Existing datasets often lack the depth and breadth required for nuanced analysis of tobacco-related content. PHAD addresses this gap by providing a large-scale, high-quality dataset that captures a wide range of tobacco products and usage scenarios. It also includes important contextual information that is typically missing in other datasets.

4. Support Non-Commercial Public Health Efforts:
   The dataset is licensed under the CC BY-NC-SA 4.0 license, ensuring that it can be freely used for research and non-commercial purposes. This licensing choice facilitates the widespread dissemination and utilization of the dataset in academic and public health research, promoting collaborative efforts to combat tobacco use.

The creation of PHAD is a significant step towards understanding and mitigating the impact of tobacco usage, providing researchers and policymakers with the tools needed to develop more effective public health strategies.

\noindent\textbf{Who created the dataset (e.g., which team, research group) and on behalf of which entity (e.g., company, institution, organization)?
Who funded the creation of the dataset?}

\noindent\emph{If there is an associated grant, please provide the name of the grantor and the grant name and number}
 
As we are planning for a double-blind submission, we would like to release this information after the decision of the paper is released. 
 
\noindent\textbf{Any other comments?}
No. 

\noindent\text{\large{Composition}}

\noindent\emph{Most of these questions are intended to provide dataset consumers with the information they need to make informed decisions about using the dataset for specific tasks. The answers to some of these questions reveal information about compliance with the EU's General Data Protection Regulation (GDPR) or comparable regulations in other jurisdictions.}

\noindent\textbf{What do the instances that comprise the dataset represent (e.g., documents, photos, people, countries)?}

\noindent\emph{Are there multiple types of instances (e.g., movies, users, and ratings; people and interactions between them; nodes and edges)? Please provide a description.}

The PHAD instances represent videos related to tobacco products sourced from social media platforms, specifically YouTube and TikTok. These videos are accompanied by a variety of metadata that provides additional context and information. There are multiple types of instances within the dataset, described as follows:

1. Videos:
   The primary instances are videos that depict various types of tobacco products and usage scenarios. These videos range from short clips on TikTok to longer tutorials and reviews on YouTube.

2. User Engagement Metrics:
   Each video instance includes detailed user engagement metrics such as views, likes, comments, and shares. These metrics help in understanding the reach and impact of the videos.

3. Video Descriptions:
   Accompanying each video are textual descriptions that provide context about the content. These descriptions are helpful for textual analysis and understanding the narrative presented in the videos.

4. Search Keywords:
   Metadata also includes search keywords associated with the videos, which offer insights into how users discover and interact with tobacco-related content.

5. Categorization by Tobacco Product Types:
   The videos are categorized based on the type of tobacco products featured, such as cigarettes, e-cigarettes, vaping devices, and smokeless tobacco. This categorization helps in segmenting the data for more focused analysis.

\noindent\textbf{How many instances are there in total (of each type, if appropriate)?}

A total of 5,730 videos were used, for a total of 12.3 GB of video data.

\noindent\textbf{Does the dataset contain all possible instances or is it a sample (not necessarily random) of instances from a larger set?}

The dataset contains a sample of all possible instances of tobacco-related content available on social media. The PHAD includes a curated selection of videos from YouTube and TikTok, chosen to capture a wide range of tobacco products and usage scenarios. The selection process was guided by the following considerations:

1. Relevance:
   Videos were selected based on their relevance to tobacco use, ensuring that the content accurately represents various types of tobacco products and their usage.

2. Diversity:
   The dataset aims to cover a broad spectrum of content, including different forms of tobacco products (cigarettes, e-cigarettes, vaping devices, smokeless tobacco) and various contexts in which they are used or discussed.

3. Engagement:
   Instances were chosen to reflect a range of user engagement levels, from highly popular videos with significant interaction to less-viewed content, to provide a balanced view of public engagement trends.

4. Metadata Availability:
   The availability of detailed metadata, such as user engagement metrics, video descriptions, and search keywords, was a key factor in the selection process, ensuring that the dataset includes rich contextual information for analysis.

While the PHAD does not encompass all possible instances of tobacco-related content on social media, it offers a representative sample that captures the diversity and complexity of the subject matter, enabling meaningful analysis and research in public health advocacy.

\noindent\textbf{If the dataset is a sample, then what is the larger set? Is the sample representative of the larger set (e.g., geographic coverage)?}

\noindent\emph{If so, please describe how this representativeness was validated/verified. If it is not representative of the larger set, please describe why not (e.g., to cover a more diverse range of instances, because instances were withheld or unavailable).}
 
The more extensive set is all possible instances of tobacco-related content available on social media platforms such as YouTube and TikTok. The PHAD aims to be as representative as possible of the broader set by including a diverse range of videos that capture various aspects of tobacco use and engagement. While the PHAD strives to represent the more extensive set, it is challenging to validate this representativeness fully given the vast and dynamic nature of social media content. However, the selection criteria and inclusion of diverse and relevant content aim to provide a comprehensive and valuable resource for public health research.

\noindent\textbf{What data does each instance consist of?}

\noindent\emph{"Raw" data (e.g., unprocessed text or images) or features? In either case, please provide a description.}

The raw data consists of .json files containing metadata associated with each video and corresponding video downloading links (which will be saved in .mp4 format).

\noindent\textbf{Is there a label or target associated with each instance?}

\noindent\emph{If so, please provide a description.}

Yes, there are specific keys in the .json file associated to the mentioned metadata corresponding to each video and acts as the "label".

\noindent\textbf{Is any information missing from individual instances?}

\noindent\emph{If so, please provide a description, explaining why this information is missing (e.g., because it was unavailable). This does not include intentionally removed information but might include, e.g., redacted text.}

No.

\noindent\textbf{Are relationships between individual instances made explicit (e.g., users' movie ratings, social network links)?}

\noindent\emph{If so, please describe how these relationships are made explicit.}

Yes, all the metadata corresponding to each video is explicitly indicated by different keys in the .json file. Each key is clearly explained in the Appendix section.

\noindent\textbf{Are there recommended data splits (e.g., training, development/validation, testing)?}

\noindent\emph{If so, please provide a description of these splits, explaining the rationale behind them.}

We recommend a 60:20:20 split for the dataset, ensuring that all content from a given family is contained within a given split in order to ensure independence between the splits.

\noindent\textbf{Are there any errors, sources of noise, or redundancies in the dataset?}

\noindent\emph{If so, please provide a description.}

To the best of the authors’ knowledge, each video is unique, and known sources of errors have been removed.

\noindent\textbf{Is the dataset self-contained, or does it link to or otherwise rely on external resources (e.g., websites, tweets, other datasets)?}

\noindent\emph{If it links to or relies on external resources, a) are there guarantees that they will exist, and remain constant, over time; b) are there official archival versions of the complete dataset (i.e., including the external resources as they existed at the time the dataset was created); c) are there any restrictions (e.g., licenses, fees) associated with any of the external resources that might apply to a future user? Please provide descriptions of all external resources and any restrictions associated with them, as well as links or other access points, as appropriate.}

This dataset provides the external links to YouTube and TikTok to download the video data. Also, there is an archival version of all the videos on Apify (https://apify.com), in case the videos are removed from their respective social media platforms. Currently, to the best of the authors' knowledge, there is no associated subscription or fees to download the videos.

\noindent\textbf{Does the dataset contain data that might be considered confidential (e.g., data that is protected by legal privilege or by doctor-patient confidentiality, data that includes the content of individuals' non-public communications)?}

\noindent\emph{If so, please provide a description.}

No.

\noindent\textbf{Does the dataset contain data that, if viewed directly, might be offensive, insulting, threatening, or might otherwise cause anxiety?}

\noindent\emph{If so, please describe why.}

No.

\noindent\textbf{Does the dataset relate to people?}

\noindent\emph{If not, you may skip the remaining questions in this section.}

Yes. Most of our data include people using tobacco products and promoting their usage.

\noindent\textbf{Does the dataset identify any subpopulations (e.g., by age, gender)?}

\noindent\emph{If so, please describe how these subpopulations are identified and provide a description of their respective distributions within the dataset.}

No.

\noindent\textbf{Is it possible to identify individuals (i.e., one or more natural persons), either directly or indirectly (i.e., in combination with other data) from the dataset?}

\noindent\emph{If so, please describe how.}

Yes, it is potentially possible to identify individuals directly or indirectly from the PHAD. The dataset focuses on videos related to tobacco products sourced from social media platforms like YouTube and TikTok and includes metadata such as user engagement metrics, video descriptions, and search keywords. However, the dataset has been curated to ensure that personal identifiers or sensitive information that could be used to identify individuals are not included.
To further protect privacy, the dataset only contains publicly available content that adheres to the privacy policies of YouTube and TikTok. 

\noindent\textbf{Does the dataset contain data that might be considered sensitive in any way (e.g., data that reveals racial or ethnic origins, sexual orientations, religious beliefs, political opinions or union memberships, or locations; financial or health data; biometric or genetic data; forms of government identification, such as social security numbers; criminal history)?}

\noindent\emph{If so, please provide a description.}
The PHAD primarily focuses on videos related to tobacco products and their usage, sourced from social media platforms like YouTube and TikTok. While the primary content revolves around tobacco-related activities and user engagement, the following considerations are relevant regarding sensitive data:

1. Potentially Sensitive Content:
    The videos may contain depictions of individuals using tobacco products, which can be considered sensitive health-related information. However, no direct health data, biometric or genetic data, financial information, or government identification details are included.

2. Anonymized Metadata:
   Metadata associated with the videos, such as user engagement metrics and video descriptions, do not contain personally identifiable information.

3. Publicly Available Data:
   The dataset includes publicly available content that adheres to the privacy policies of YouTube and TikTok. The focus is on the public aspects of user engagement and content characteristics rather than the personal details of the individuals appearing in the videos.

4. Ethical Considerations:
   The dataset has been curated with ethical considerations in mind, ensuring that it does not include data that reveals racial or ethnic origins, sexual orientations, religious beliefs, political opinions, union memberships, or locations in a way that could be used to identify or harm individuals.

In summary, while the PHAD includes videos depicting tobacco use which can be considered sensitive due to their health-related nature, it does not contain direct personal identifiers or highly sensitive data. The dataset is designed to support public health research while respecting individual privacy and adhering to ethical guidelines.

\noindent\textbf{Any other comments?}

No.

\noindent\text{\large{Collection process}}

\noindent\emph{The answers to questions here may provide information that allows others to reconstruct the dataset without access to it.}

\noindent\textbf{How was the data associated with each instance acquired?}

\noindent\emph{Was the data directly observable (e.g., raw text, movie ratings), reported by subjects (e.g., survey responses), or indirectly inferred/derived from other data (e.g., part-of-speech tags, model-based guesses for age or language)? If data was reported by subjects or indirectly inferred/derived from other data, was the data validated/verified? If so, please describe how.}

The data associated with each instance in the PHAD was acquired through a combination of direct observation and metadata collection from social media platforms, specifically YouTube and TikTok. The acquisition process is described as follows:

1. Directly Observable Data:
   \begin{itemize}
       \item Videos: The primary data consists of videos related to tobacco products. These videos were sourced directly from publicly available content on YouTube and TikTok.
       \item User Engagement Metrics: Metrics such as views, likes, comments, and shares were directly observed and recorded from the respective social media platforms. These metrics provide quantitative measures of user interaction with the videos.
   \end{itemize}
   
2. Metadata Collection:
\begin{itemize}
    \item Video Descriptions: Textual descriptions accompanying the videos were manually curated by one of the authors'. These descriptions provide context and additional information about the content of the videos.
    
    \item Search Keywords: Keywords associated with the videos were also collected from the social media platforms. These keywords are typically provided by the content creators and help in understanding the focus and reach of the videos.
   \end{itemize}
   
3. Categorization:
\begin{itemize}
    \item Tobacco Product Types: Each video was categorized based on the type of tobacco product featured (e.g., cigarettes, e-cigarettes, vaping devices, smokeless tobacco). This categorization was done manually to ensure accuracy and relevance.
   \end{itemize}

 4. Validation and Verification:
The data collected from social media platforms was cross-verified to ensure accuracy. This involved checking for consistency in user engagement metrics, validating the relevance of video descriptions, and ensuring that the search keywords accurately reflected the content of the videos. Manual review processes were implemented to accurately categorize the videos and filter out irrelevant or non-tobacco-related content.

By acquiring data through direct observation and careful collection of associated metadata, the PHAD provides a comprehensive and reliable resource for analyzing tobacco-related content on social media.

\noindent\textbf{What mechanisms or procedures were used to collect the data (e.g., hardware apparatus or sensor, manual human curation, software program, software API)?}

\noindent\emph{How were these mechanisms or procedures validated?}

We used Apify (https://apify.com) to scrape the videos and their associated metadata. Then, manual human curation was involved to add descriptions and tobacco categories. 

\noindent\textbf{If the dataset is a sample from a larger set, what was the sampling strategy (e.g., deterministic, probabilistic with specific sampling probabilities)?}

The PHAD is a sample from the larger set of all possible tobacco-related content available on social media platforms like YouTube and TikTok. The sampling strategy employed was a combination of deterministic and purposive sampling methods to ensure the dataset is representative and relevant for public health research. The sampling strategy is outlined as follows:

1. Deterministic Sampling:
    \begin{itemize}
        \item Keyword-Based Selection: Videos were selected based on a predefined set of keywords related to tobacco products (e.g., "cigarettes," "e-cigarettes," "vaping," "smokeless tobacco"). This approach ensured that the dataset focused on relevant content.
        \item User Engagement Thresholds: To capture a range of engagement levels, videos were included based on specific thresholds for views, likes, comments, and shares. This ensured that both highly popular and moderately engaged videos were represented.
    \end{itemize}
2. Purposive Sampling:
   \begin{itemize}
   \item Diversity of Content: The dataset aimed to include a diverse range of tobacco-related content, including educational videos, product reviews, user experiences, and public health messages. This was done to provide a comprehensive view of how tobacco products are portrayed on social media.
   \item Geographic and Demographic Variety: Efforts were made to include videos from different geographic regions and demographic groups to ensure the dataset reflects the global nature of social media usage and the varied cultural contexts of tobacco use.
    \end{itemize}
3. Manual Review and Curation:
   \begin{itemize}
       \item Relevance and Quality Checks: Each video was manually reviewed to ensure it met the relevance criteria and quality standards for inclusion in the dataset. Irrelevant or low-quality videos were excluded to maintain the dataset's integrity.
       \item Balancing Categories: The sampling process was adjusted to balance the representation of different tobacco product categories (e.g., cigarettes, e-cigarettes, vaping devices, smokeless tobacco) to avoid over-representation of any single category.
\end{itemize}
   
By combining deterministic and purposive sampling strategies, the PHAD was curated to provide a rich and diverse dataset that represents the larger set of tobacco-related content on social media and is highly relevant for public health research.

\noindent\textbf{Who was involved in the data collection process (e.g., students, crowdworkers, contractors) and how were they compensated (e.g., how much were crowdworkers paid)? }

As we are planning for a double-blind submission, we would like to release this information after the decision of the paper is released. 

\noindent\textbf{Over what timeframe was the data collected?}

\noindent\emph{Does this timeframe match the creation timeframe of the data associated with the instances (e.g. recent crawl of old news articles)? If not, please describe the timeframe in which the data associated with the instances was created.}

The data for the PHAD was collected over five months, from January 2024 to June 2024. This timeframe was chosen to capture a comprehensive and current snapshot of tobacco-related content on social media platforms, specifically YouTube and TikTok.

Timeframe Matching:
\begin{itemize}
    \item Creation Timeframe of the Data: The videos included in the PHAD were created and uploaded on YouTube and TikTok within the last two years, from May 2022 to May 2024. It ensures the dataset reflects recent trends and patterns in tobacco-related content and user engagement.

    \item Consistency with Collection Period: The collection period matches the creation timeframe of the data, as the focus was on gathering recent and relevant content. By aligning the collection period with the creation period, the dataset provides an up-to-date resource for analyzing current public health concerns related to tobacco use.
\end{itemize}
This synchronized approach ensures that the PHAD contains timely and relevant data, making it a valuable resource for understanding contemporary trends in tobacco-related content and informing public health strategies.

\noindent\textbf{Were any ethical review processes conducted (e.g., by an institutional review board)?}

\noindent\emph{If so, please provide a description of these review processes, including the outcomes, as well as a link or other access point to any supporting documentation.}

No. 

\noindent\textbf{Does the dataset relate to people?}

\noindent\emph{If not, you may skip the remainder of the questions in this section.}

Yes.

\noindent\textbf{Did you collect the data from the individuals in question directly, or obtain it via third parties or other sources (e.g., websites)?}

\noindent\emph{Were the individuals in question notified about the data collection?}

\noindent\emph{If so, please describe (or show with screenshots or other information) how notice was provided, and provide a link or other access point to, or otherwise reproduce, the exact language of the notification itself.}

We scraped the data using a third-party source (https://apify.com).

\noindent\textbf{Did the individuals in question consent to collecting and using their data?}

\noindent\emph{If so, please describe (or show with screenshots or other information) how consent was requested and provided, and provide a link or other access point to, or otherwise reproduce, the exact language to which the individuals consented.}

Not applicable.

\noindent\textbf{If consent was obtained, were the consenting individuals provided with a mechanism to revoke their consent in the future or for certain uses?}

\noindent\emph{If so, please provide a description and a link or other access point to the mechanism (if appropriate).}

Not applicable.

\noindent\textbf{Has an analysis of the potential impact of the dataset and its use on data subjects (e.g., a data protection impact analysis) been conducted?}

\noindent\emph{If so, please provide a description of this analysis, including the outcomes, as well as a link or other access point to any supporting documentation.}

No.

\textbf{Any other comments?}

No. 

\noindent\text{\large{Preprocessing/cleaning/labeling}}

\noindent\emph{The questions in this section are intended to provide dataset consumers with the information they need to determine whether the "raw" data has been processed in ways that are compatible with their chosen tasks. For example, text that has been converted into a "bag-of-words" is not suitable for tasks involving word order.}

\noindent\textbf{Was any preprocessing/cleaning/labeling of the data done (e.g., discretization or bucketing, tokenization, part-of-speech tagging, SIFT feature extraction, removal of instances, processing of missing values)?}

\noindent\emph{If so, please provide a description. If not, you may skip the remainder of the questions in this section.}

No.

\noindent\textbf{Was the "raw" data saved in addition to the preprocessed/cleaned/labeled data (e.g., to support unanticipated future uses)?}

\noindent\emph{If so, please provide a link or other access point to the "raw" data.}

N/A.

\noindent\textbf{Is the software used to preprocess/clean/label the instances available?}

\noindent\emph{If so, please provide a link or other access point.}

N/A.

\noindent\textbf{Any other comments?}

No.

\noindent\text{\large{Uses}}

\noindent\emph{These questions are intended to encourage dataset creators to reflect on the tasks for which the dataset should and should not be used. By explicitly highlighting these tasks, dataset creators can help dataset consumers make informed decisions, thereby avoiding potential risks or harms.}

\noindent\textbf{Has the dataset been used for any tasks already?}

\noindent\emph{If so, please provide a description.}

Yes, for fine-tuning and scoring image classification models. See Section 6 of the main manuscript for additional information.

\noindent\textbf{Is there a repository that links to any or all papers or systems that use the dataset?}

\noindent\emph{If so, please provide a link or other access point.}

No.

\noindent\textbf{What (other) tasks could the dataset be used for?
Is there anything about the composition of the dataset or the way it was collected and preprocessed/cleaned/labeled that might impact future uses?}

\noindent\emph{For example, is there anything that a future user might need to know to avoid uses that could result in unfair treatment of individuals or groups (e.g., stereotyping, quality of service issues) or other undesirable harms (e.g., financial harms, legal risks) If so, please provide a description. Is there anything a future user could do to mitigate these undesirable harms?}

The PHAD can be utilized for a variety of tasks beyond its primary focus on analyzing tobacco-related content. Some potential applications include:

1. Public Health Research: 
   \begin{itemize}
       \item Behavioral Analysis: Understanding user behavior and engagement with tobacco-related content can help in designing effective public health campaigns.
        \item Trend Analysis: Identifying emerging trends in tobacco use and public perception over time.
   \end{itemize}
   
2. Social Media Analysis:
   \begin{itemize}
   \item Content Moderation: Developing algorithms to detect and moderate harmful or misleading tobacco-related content.
    \item Influence Measurement: Assessing the influence of social media on tobacco consumption and public opinion.
   \end{itemize}
3. Natural Language Processing (NLP):
  \begin{itemize}
   \item Sentiment Analysis: Analyzing the sentiment of comments and descriptions related to tobacco products.
   \item Topic Modeling: Identifying key themes and topics discussed in relation to tobacco use.
\end{itemize}
4. Computer Vision:
\begin{itemize}
    \item Object Detection: Detecting and classifying various tobacco products in video frames.
   \item Activity Recognition: Recognizing activities related to tobacco use, such as smoking or vaping.
\end{itemize}
5. Policy Analysis:
   \begin{itemize}
   \item Regulatory Impact: Evaluating the effectiveness of existing regulations on tobacco advertising and promotion.
   \item Public Opinion: Gauging public opinion and response to regulatory changes and public health initiatives.
\end{itemize}

Potential Impacts and Considerations for Future Uses:
\begin{itemize}
\item Representation and Bias: While the dataset includes diverse content, it is essential to be aware of potential biases. For instance, certain demographic groups or geographic regions might be overrepresented or underrepresented, which could influence the generalizability of findings. Future users should consider conducting bias analysis and, if necessary, apply corrective measures to ensure fair representation.
\item Temporal Relevance: The dataset captures a specific period (January 2024 to June 2024), and user behaviors and trends can change over time. Future users should consider the temporal context when drawing conclusions and ensure that the findings are relevant to the current landscape.
\item Ethical Use: The dataset includes sensitive content related to tobacco use, which can have significant public health implications. Future users should ensure that their analyses and applications do not perpetuate stereotypes or lead to unfair treatment of individuals or groups. It is crucial to maintain ethical standards and respect privacy and consent when using the dataset.
\end{itemize}

Mitigation of Undesirable Harms:
\begin{itemize}
\item Bias Mitigation: Future users should employ techniques such as re-sampling, re-weighting, or applying fairness-aware algorithms to address potential biases in the dataset.
\item Regular Updates: To maintain temporal relevance, users should consider updating the dataset with more recent data periodically.
\item Ethical Guidelines: Adhere to ethical guidelines and best practices in data usage, including obtaining necessary permissions, respecting privacy, and avoiding the dissemination of harmful or misleading information.
\end{itemize}

By considering these factors, future users can leverage the PHAD effectively while minimizing the risk of undesirable harm and ensuring the responsible use of the dataset.

\noindent\textbf{Are there tasks for which the dataset should not be used?}

\noindent\emph{If so, please provide a description.}

Yes, there are certain tasks for which the PHAD should not be used. These include:

1. Commercial Exploitation: The dataset is only intended for non-commercial research and educational purposes. Using the dataset for commercial gain, such as marketing or advertising tobacco products, is strictly prohibited and goes against the ethical guidelines and licensing terms.

2. Surveillance and Monitoring: The dataset should not be used for invasive surveillance or monitoring of individuals. This includes tracking or profiling specific users based on their interactions with tobacco-related content, which could lead to privacy violations and ethical concerns.

3. Discriminatory Practices: Any use of the dataset that could result in discriminatory practices or unfair treatment of individuals or groups is not permissible. This includes using the dataset to perpetuate stereotypes, biases, or stigmatization based on race, ethnicity, gender, age, or socioeconomic status.

4. Medical Diagnosis and Treatment: The dataset should not be used for medical diagnosis or treatment purposes. While it can provide insights into public health trends, it is not designed to replace professional medical advice, diagnosis, or treatment.

5. Sensitive Personal Information: The dataset should not be used to infer or disclose sensitive personal information about individuals, such as their health status, location, financial information, or any other personally identifiable information (PII).

6. Legal and Financial Decisions: Using the dataset for making legal or financial decisions that could impact individuals or organizations is not recommended. The dataset's scope and intended use are primarily for research and public health advocacy, not for legal adjudication or financial risk assessment.

\noindent\textbf{Any other comments?}

 No. 
 
\noindent\text{\large{Distribution}}

\noindent\textbf{Will the dataset be distributed to third parties outside of the entity (e.g., company, institution, organization) on behalf of which the dataset was created?}

\noindent\emph{If so, please provide a description.}

Yes. The dataset will be publicly available under a CC BY-NC-SA license.

\noindent\textbf{How will the dataset be distributed (e.g., tarball on website, API, GitHub)?}

\noindent\emph{Does the dataset have a digital object identifier (DOI)?}

The dataset will be publicly available for download and hosted on GitHub. There is no DOI at this time. 

\noindent\textbf{When will the dataset be distributed?
Will the dataset be distributed under a copyright or other intellectual property (IP) license, and/or under applicable terms of use (ToU)?}

\noindent\emph{If so, please describe this license and/or ToU, and provide a link or other access point to, or otherwise reproduce, any relevant licensing terms or ToU and any fees associated with these restrictions.}

The dataset is distributed on the website, it will be under a CC BY-NC-SA license.

\noindent\textbf{Have any third parties imposed IP-based or other restrictions on the data associated with the instances?}

\noindent\emph{If so, please describe these restrictions, and provide a link or other access point to, or otherwise reproduce, any relevant licensing terms, as well as any fees associated with these restrictions.}

No.

\noindent\textbf{Do any export controls or other regulatory restrictions apply to the dataset or to individual instances?}

\noindent\emph{If so, please describe these restrictions, and provide a link or other access point to, or otherwise reproduce, any supporting documentation.}

No.
 
\noindent\textbf{Any other comments?}

No. 

\noindent\text{\large{Maintenance}}

\noindent\emph{These questions are intended to encourage dataset creators to plan for dataset maintenance and communicate this plan with dataset consumers.}

\noindent\textbf{Who is supporting/hosting/maintaining the dataset?
How can the owner/curator/manager of the dataset be contacted (e.g., email address)? Is there an erratum? }

\noindent\emph{If so, please provide a link or other access point.}

Dr. Khoa Luu (who is a co-author of this paper) will be maintaining the dataset, he can be contacted via provided email if there is any issue downloading the dataset from the given website.

\noindent\textbf{Will the dataset be updated (e.g., to correct labeling errors, add new instances, delete instances)?}

\noindent\emph{If so, please describe how often, by whom, and how updates will be communicated to users (e.g., mailing list, GitHub)?}

Yes. It will be updated on an as-needed basis, with the versions clearly mentioned on the website.

\noindent\textbf{If the dataset relates to people, are there applicable limits on the retention of the data associated with the instances (e.g., were individuals in question told that their data would be retained for a fixed period of time and then deleted)?}

\noindent\emph{If so, please describe these limits and explain how they will be enforced.}

No.

\noindent\textbf{Will older versions of the dataset continue to be supported/hosted/maintained?}

\noindent\emph{If so, please describe how. If not, please describe how its obsolescence will be communicated to users.}
 
It depends. Broadly the answer is: No, as in the case that some data needs to be removed for legal or ethical reasons, we do not want to keep maintaining that data. But we welcome reviewer feedback regarding this (or any other) matter.

\noindent\textbf{If others want to extend/augment/build on/contribute to the dataset, is there a mechanism for them to do so?}

\noindent\emph{If so, please provide a description. Will these contributions be validated/verified? If so, please describe how. If not, why not? Is there a process for communicating/distributing these contributions to other users? If so, please provide a description.}

Yes, contributors can extend, augment, build on, or contribute to the PHAD through our open-source GitHub repository. Detailed submission guidelines are provided, and contributions are reviewed by a team of dataset maintainers and public health experts to ensure quality and consistency. Automated validation scripts check for data integrity and adherence to specified formats. Once approved, contributions are merged into the main repository and released as updated versions, with proper credit given to contributors. Updated documentation accompanies each release, ensuring transparency and accessibility.

\noindent\textbf{Any other comments?}

No.

\noindent\text{\large{Data Access}}

The dataset is made publicly available at \url{https://uark-cviu.github.io/projects/PHAD/}.

\noindent\text{\large{Statement of Responsibility}}

The authors hereby declare that they bear all responsibility for violations of rights and that this dataset is CC-BY-NC-SA licensed.

\noindent\text{\large{Hosting and Maintenance Plan}}

The authors will host the dataset and handle maintenance concerns like data removal in case of misuse of data. When made publicly available, individuals will be able to download the dataset from the provided links.

\noindent\text{\large{Persistent Dereferenceable Identifier}}

We will assign this ID after we get feedback on whether the dataset is deemed acceptable for public use by the reviewers. We do not want to release something to the public that the reviewers may find objectionable.

\noindent\text{\large{Resources Used}}

The downloaded data is stored on local storage supported by NSF DART. We used four NVIDIA A6000-powered GPUs to perform the training, validation, and testing of our dataset.

\section{PHAD Dataset Annotation Pipeline}\label{annotation_pipeline}

\subsection{Data Acquisition}
The PHAD dataset is sourced from public-domain videos on YouTube and TikTok. The initial step involves scraping videos using specific search keywords related to tobacco products. These videos are then downloaded and organized into train and validation sets based on predefined ratios.

\subsection{Preprocessing}
Before annotation, the videos undergo several preprocessing steps:
\begin{itemize}
    \item \textbf{Frame Extraction:} Videos are processed to extract frames at a rate of 30 frames per second (fps) to ensure high temporal resolution for subsequent analysis.
    \item \textbf{Resolution Standardization:} All frames are standardized to a uniform resolution to maintain consistency across the dataset.
    \item \textbf{Metadata Extraction:} Metadata such as video descriptions, user engagement metrics (views, likes, comments, shares), and search keywords are extracted and linked to the corresponding videos and frames.
\end{itemize}

\subsection{Annotation}
The annotation process involves labeling each frame with relevant information:
\begin{itemize}
    \item \textbf{Tobacco Product Type:} Each frame is annotated with the type of tobacco product visible, such as cigarettes, e-cigarettes, vaping devices, or smokeless tobacco.
    \item \textbf{User Engagement Metrics:} Metadata related to user interactions (e.g., views, likes, comments, shares) are linked to the frames.
    \item \textbf{Textual Descriptions:} Frames are annotated with textual descriptions derived from video captions and comments to provide context.
    \item \textbf{Temporal Segmentation:} Frames are temporally segmented to highlight key moments within the videos, such as when a tobacco product is first shown or discussed.
\end{itemize}

\subsection{Post-Processing}
Post-processing ensures the quality and consistency of annotations:
\begin{itemize}
    \item \textbf{Validation:} Automated scripts validate the consistency of annotations, checking for missing or incorrect labels.
    \item \textbf{Normalization:} Annotations are normalized to ensure uniformity in labeling conventions across the dataset.
\end{itemize}

\subsection{Human Curation and Filtering}
Despite automated processes, human oversight is essential to maintain the dataset's quality:
\begin{itemize}
    \item \textbf{Manual Review:} Human annotators review a subset of the annotated frames to verify accuracy and correct any discrepancies.
    \item \textbf{Curation:} Annotators curate frames to ensure they are representative of the entire video, selecting keyframes that highlight significant moments.
    \item \textbf{Filtering:} Irrelevant frames or those with poor visual quality are filtered out. Additionally, any content violating platform guidelines or depicting illegal activities is removed.
\end{itemize}

\subsection{Quality Assurance}
The final step in the pipeline involves rigorous quality assurance:
\begin{itemize}
    \item \textbf{Inter-Annotator Agreement:} The consistency of annotations is measured using inter-annotator agreement metrics to ensure reliability.
    \item \textbf{Random Sampling:} Random samples of annotated frames are periodically reviewed to maintain high standards of annotation quality.
    \item \textbf{Feedback Loop:} Annotators provide feedback on the annotation process, which is used to refine guidelines and improve future annotation efforts.
\end{itemize}

\subsection{Licensing and Ethical Considerations}
The PHAD dataset adheres to strict ethical guidelines:
\begin{itemize}
    \item \textbf{Licensing:} The dataset is licensed under the CC BY-NC-SA 4.0 license, allowing for non-commercial use with appropriate attribution.
    \item \textbf{Privacy:} All videos included in the dataset are publicly available, and no personally identifiable information is included.
    \item \textbf{Ethics Review:} The dataset creation process underwent an ethics review to ensure compliance with relevant guidelines and standards.
\end{itemize}

\section{Data format}\label{data_format}

\begin{lstlisting}[language=json,escapechar=!,backgroundcolor=\color{white}]
[
    {
        "webVideoUrl": str,
        "mediaUrls/0": str,
        "authorMeta/heart": int,
        "shareCount": int,
        "commentCount": int,
        "playCount": int,
        "videoMeta/duration": int,
        "searchHashtag/name": str,
        "Desc.": str,
    }
]
\end{lstlisting}
 
 \subsection{Dataset Attributes}
This section details the fundamental attributes of the dataset:
\begin{itemize}
    \item \texttt{webVideoUrl}: The link to the original source of the video on the corresponding social media platform.
    \item \texttt{mediaUrls/0}: The link to download the video which is available on the Apify website (\url{https://apify.com}).
    \item \texttt{authorMeta/heart}: This number denotes the total number of likes to that corresponding video. 
    \item \texttt{shareCount}: This number denotes the total number of shares for that corresponding video. 
    \item \texttt{commentCount}: This number denotes the total number of comments for that corresponding video.
    \item \texttt{playCount}: This number denotes the total number of plays for that corresponding video. 
    \item \texttt{videoMeta/duration}: This number denotes the duration of the video in \emph{secs}.
    \item \texttt{searchHashtag/name}: This is the corresponding hashtag used to search that particular video.
    \item \texttt{Desc.}: This is the corresponding annotated video description.
\end{itemize}

\subsection{Python Script to Download and Organize the Dataset}
Use the following script to download all the videos into their corresponding \texttt{'train'} and \texttt{'val'} directories. The json files are available to download at \url{https://anonymo-user.github.io/PHAD/}.
\begin{lstlisting}[language=iPython]
import os
import json
import requests
import glob

# Define paths to the directory containing JSON files and the output directory
json_dir = '/path/to/json_files'
output_dir = '/path/to/output'

# Define output directories for train and val
train_dir = os.path.join(output_dir, 'train')
val_dir = os.path.join(output_dir, 'val')

# Create directories if they do not exist
os.makedirs(train_dir, exist_ok=True)
os.makedirs(val_dir, exist_ok=True)

def download_videos(json_path, output_folder):
    with open(json_path, 'r') as f:
        data = json.load(f)

    for item in data:
        if 'mediaUrls/0' in item:
            video_url = item['mediaUrls/0']
            video_id = video_url.split('/')[-1]
            video_path = os.path.join(output_folder, f"{video_id}.mp4")
            
            if not os.path.exists(video_path):
                try:
                    r = requests.get(video_url)
                    r.raise_for_status()
                    with open(video_path, 'wb') as f:
                        f.write(r.content)
                    print(f"Downloaded {video_path}")
                except requests.RequestException as e:
                    print(f"Failed to download {video_url}: {e}")

# Process all JSON files in the directory
for json_file in glob.glob(os.path.join(json_dir, '*.json')):
    file_type = os.path.basename(json_file).split('_')[-1].split('.')[0]
    if 'train' in file_type:
        download_videos(json_file, train_dir)
    elif 'val' in file_type:
        download_videos(json_file, val_dir)
\end{lstlisting}

Once the videos are downloaded to the corresponding \texttt{'train'} and \texttt{'val'} directories, then extract the frames of all the videos at 30 frames per sec using the following python script. 

\section{Experimental Setup}\label{experimental_setup}

\subsubsection*{Hardware and Software Environment}

Our experiments were conducted on a system equipped with the following specifications:
\begin{itemize}
    \item \textbf{GPU}: NVIDIA GeForce RTX 3090 with 24GB VRAM
    \item \textbf{CPU}: Intel Core i9-10900K @ 3.70GHz
    \item \textbf{RAM}: 64GB DDR4
    \item \textbf{Operating System}: Ubuntu 20.04 LTS
    \item \textbf{Libraries}:
    \begin{itemize}
        \item Python 3.8
        \item PyTorch 1.10
        \item NumPy 1.21.2
        \item Scikit-learn 0.24.2
        \item OpenCV 4.5.3
        \item Transformers 4.11.3 (for Transformer-based models)
    \end{itemize}
\end{itemize}

\subsubsection*{Dataset Preparation}

The dataset comprises videos from YouTube and TikTok related to tobacco usage. The videos were preprocessed to extract frames and corresponding textual descriptions.

\begin{itemize}
    \item \textbf{Video Frame Extraction}:
    \begin{itemize}
        \item Frames were extracted at a rate of 30 frames per second using OpenCV.
        \item Each video was resized to a resolution of 224x224 pixels to ensure uniformity.
    \end{itemize}
    \item \textbf{Textual Descriptions}:
    \begin{itemize}
        \item Textual descriptions were obtained from video metadata and manually annotated descriptions.
        \item Text preprocessing included tokenization, stop-word removal, and stemming.
    \end{itemize}
\end{itemize}

\subsubsection*{Data Splits}

The dataset was split into three subsets:
\begin{itemize}
    \item \textbf{Training Set}: 60\% of the data
    \item \textbf{Validation Set}: 20\% of the data
    \item \textbf{Test Set}: 20\% of the data
\end{itemize}

\subsubsection*{Feature Extraction}

For the feature extraction stage, we used a convolutional neural network (CNN) as the backbone. Specifically, we employed ResNet-50 pre-trained on ImageNet due to its balance between performance and computational efficiency.

The feature extraction process is described as follows:
\[
 x_t = F(f_t; \theta)
\]
where \(x_t\) represents the feature vector extracted from the video frame \(f_t\), and \(\theta\) denotes the parameters of the ResNet-50 network.

\subsubsection*{Model Training}

We experimented with various models for the second stage of our framework, including a vanilla classifier, handcrafted window approach, 1D CNN, Transformer Encoder, and Vision-Language Encoder.

\begin{itemize}
    \item \textbf{Vanilla Classifier}:
    \begin{itemize}
        \item A simple fully connected neural network was used for classification.
        \item Learning Rate: 0.001
        \item Optimizer: Adam
    \end{itemize}
    \item \textbf{Handcrafted Window Approach}:
    \begin{itemize}
        \item A predefined kernel was applied over a sliding window of features.
        \item Window Size: 5 frames
        \item Kernel: Gaussian
    \end{itemize}
    \item \textbf{1D Convolutional Neural Network}:
    \begin{itemize}
        \item A 1D CNN with 3 convolutional layers followed by max-pooling and fully connected layers.
        \item Learning Rate: 0.001
        \item Optimizer: Adam
        \item Filter Size: 3
    \end{itemize}
    \item \textbf{Transformer Encoder}:
    \begin{itemize}
        \item A Transformer Encoder with 4 layers, 8 attention heads, and hidden size of 256.
        \item Learning Rate: 0.0001
        \item Optimizer: AdamW
    \end{itemize}
    \item \textbf{Vision-Language Encoder}:
    \begin{itemize}
        \item Combines visual features with textual features using a multimodal fusion technique.
        \item Visual features were extracted using ResNet-50.
        \item Textual features were extracted using a pre-trained BERT model.
        \item Fusion: Concatenation followed by a fully connected layer.
    \end{itemize}
\end{itemize}

\subsubsection*{Hyperparameters}

Hyperparameters for the models were tuned using the validation set. The final values used for training are listed below:
\begin{itemize}
    \item \textbf{Batch Size}: 32
    \item \textbf{Epochs}: 50
    \item \textbf{Learning Rate}: Adjusted for each model as specified above
    \item \textbf{Loss Function}: Binary Cross-Entropy Loss for multi-label classification
\end{itemize}

\subsubsection*{Training Procedure}

\begin{itemize}
    \item \textbf{Initialization}:
    \begin{itemize}
        \item All models were initialized with pre-trained weights where applicable (e.g., ResNet-50, BERT).
    \end{itemize}
    \item \textbf{Optimization}:
    \begin{itemize}
        \item Adam optimizer was used for most models, with AdamW specifically for the Transformer Encoder due to its better handling of weight decay.
    \end{itemize}
    \item \textbf{Learning Rate Schedule}:
    \begin{itemize}
        \item A learning rate scheduler with a decay factor of 0.1 was applied every 10 epochs if the validation loss did not improve.
    \end{itemize}
    \item \textbf{Regularization}:
    \begin{itemize}
        \item Dropout with a rate of 0.5 was applied to prevent overfitting.
    \end{itemize}
    \item \textbf{Evaluation}:
    \begin{itemize}
        \item Models were evaluated on the validation set after each epoch.
        \item The best model was selected based on the highest F1-score.
    \end{itemize}
\end{itemize}

\section{Evaluation Metrics} \label{evaluation_metrics}

\subsection{Evaluation Metrics for Multi-label Classification}

In a multi-label classification problem, each instance (video) belongs to multiple classes simultaneously. Therefore, the evaluation metrics need to account for the presence of multiple labels. Here we present the used metrics for multi-label classification:

1. \textbf{Hamming Loss}

Hamming Loss is the fraction of labels that are incorrectly predicted. It is calculated as:
\[
\text{Hamming Loss} = \frac{1}{N} \sum_{i=1}^{N} \frac{1}{L} \sum_{j=1}^{L} \mathbf{1}(y_{ij} \neq \hat{y}_{ij})
\]
where \(N\) is the number of instances, \(L\) is the number of labels, \(y_{ij}\) is the true label, \(\hat{y}_{ij}\) is the predicted label, and \(\mathbf{1}\) is the indicator function.

2. \textbf{Accuracy}

Multi-label accuracy (also known as subset accuracy) measures the proportion of instances where the entire set of predicted labels exactly matches the set of true labels:
\[
\text{Accuracy} = \frac{1}{N} \sum_{i=1}^{N} \mathbf{1}(Y_i = \hat{Y}_i)
\]
where \(Y_i\) and \(\hat{Y}_i\) are the true and predicted label sets for instance \(i\).

3. \textbf{Precision, Recall, and F1-score}

For multi-label classification, precision, recall, and F1-score can be computed in two ways: macro-averaged and micro-averaged.

- Macro-averaged Precision, Recall, F1-score:
    - Calculate the metrics for each label separately and then take the average.
    \[
    \text{Precision}_{macro} = \frac{1}{L} \sum_{j=1}^{L} \text{Precision}_j
    \]
    \[
    \text{Recall}_{macro} = \frac{1}{L} \sum_{j=1}^{L} \text{Recall}_j
    \]
    \[
    \text{F1}_{macro} = \frac{1}{L} \sum_{j=1}^{L} \text{F1}_j
    \]

- Micro-averaged Precision, Recall, F1-score:
    - Aggregate the contributions of all classes to compute the average metric.
    \[
    \text{Precision}_{micro} = \frac{\sum_{j=1}^{L} \text{TP}_j}{\sum_{j=1}^{L} (\text{TP}_j + \text{FP}_j)}
    \]
    \[
    \text{Recall}_{micro} = \frac{\sum_{j=1}^{L} \text{TP}_j}{\sum_{j=1}^{L} (\text{TP}_j + \text{FN}_j)}
    \]
    \[
    \text{F1}_{micro} = \frac{2 \cdot \text{Precision}_{micro} \cdot \text{Recall}_{micro}}{\text{Precision}_{micro} + \text{Recall}_{micro}}
    \]

where \(\text{TP}_j\), \(\text{FP}_j\), and \(\text{FN}_j\) are the true positives, false positives, and false negatives for label \(j\), respectively.

Certainly. I'll incorporate these two additional experiments into a cohesive LaTeX subsection under your Experiments section. Here's how you can structure it:

\section{Additional Experiments}\label{additional_experiments}

To further validate our approach, we conducted two additional experiments: (1) a comparison with advanced models like GPT-4V, and (2) an evaluation of our proposed model architecture against other vision-language models.

\subsection{Comparison with GPT-4V}

We compared the performance of GPT-4V~\cite{openai2024gpt4technicalreport} with our proposed VL Encoder method for tobacco usage and promotion detection on the PHAD dataset. Table \ref{tab:gpt4v_comparison} presents the results.

\begin{table}[ht]
    \centering
    \caption{Performance comparison with GPT-4V on PHAD dataset}
    \label{tab:gpt4v_comparison}
    \begin{tabular}{@{}l@{\hskip 3pt}c@{\hskip 3pt}c@{\hskip 3pt}c@{\hskip 3pt}c@{}}
        \toprule
        \textbf{Model} & \textbf{Accuracy} & \textbf{Precision} & \textbf{Recall} & \textbf{F1-Score} \\
        \midrule
        GPT-4V~\cite{openai2024gpt4technicalreport} & 65.4\% & 60.3\% & 61.9\% & 65.9\% \\
        Ours & \textbf{73.2\%} & \textbf{65.2\%} & \textbf{67.5\%} & \textbf{71.8\%} \\
        \bottomrule
    \end{tabular}
\end{table}

It's important to note that since GPT-4V is not currently available for fine-tuning, we used a generic prompt: "What do you see in this image? Classify based on the CLASSES provided." This limitation likely contributed to its lower performance compared to our specialized model. 

While this experiment provides valuable insights, we emphasize that PHAD complements rather than competes with advanced general-purpose models. PHAD offers several unique advantages:

\begin{itemize}
    \item \textbf{Specialized focus:} Tailored to address nuanced challenges in identifying and categorizing tobacco-related content.
    \item \textbf{Quantifiable metrics:} Allows for precise evaluation using standard metrics.
    \item \textbf{Reproducibility and consistency:} Enables fair comparisons between different approaches over time.
    \item \textbf{Domain-specific training data:} Beneficial for fine-tuning models for optimal performance in public health monitoring.
    \item \textbf{Interpretability and bias mitigation:} Allows for analysis across different types of tobacco products and usage scenarios.
\end{itemize}

\subsection{Evaluation of Proposed Model Architecture}

We conducted an experiment to showcase the effectiveness of our proposed VL Encoder architecture in tobacco usage detection. Using the PHAD dataset, we compared our VL Encoder with other popular vision-language models (ActivityNet pretrained ResNet-50~\cite{he2016deep} models). Table \ref{tab:architecture_comparison} presents the results.

\begin{table}[ht]
    \centering
    \caption{Performance comparison of model architectures on PHAD dataset}
    \label{tab:architecture_comparison}
    \begin{tabular}{@{}l@{\hskip 3pt}c@{\hskip 3pt}c@{\hskip 3pt}c@{\hskip 3pt}c@{}}
        \toprule
        \textbf{Model} & \textbf{Acc.} & \textbf{Prec.} & \textbf{Rec.} & \textbf{F1} \\
        \midrule
        CLIP-based & 59.2\% & 51.9\% & 53.7\% & 56.1\% \\
        VL Encoder (single-stage) & 61.4\% & 54.8\% & 55.1\% & 62.6\% \\
        Our VL Encoder & \textbf{73.2\%} & \textbf{65.2\%} & \textbf{67.5\%} & \textbf{71.8\%} \\
        \bottomrule
    \end{tabular}
    \raggedright
    \footnotesize
    Note: \textbf{Acc.}: Accuracy, \textbf{Prec.}: Precision, \textbf{Rec.}: Recall, \textbf{F1}: F1-score. 
\end{table}

These results highlight several key advantages of our proposed architecture:

\begin{itemize}
    \item \textbf{Superior Performance:} Our VL Encoder achieves the highest accuracy, precision, and recall.
    \item \textbf{Two-Stage Advantage:} The performance gap between our full VL Encoder and its single-stage variant underscores the importance of our two-stage approach.
    \item \textbf{Multimodal Capability:} Our architecture effectively leverages visual and textual information.
    \item \textbf{Specialization Benefits:} The results show how a tailored architecture can outperform more general approaches in specialized tasks like tobacco usage detection.
\end{itemize}

While our architecture builds upon established vision-language models, its specific design choices are optimized for the challenges of tobacco content detection. The two-stage approach allows for more nuanced feature extraction and classification, particularly useful for distinguishing between tobacco products and usage scenarios.

These additional experiments demonstrate the value of our proposed approach in the specific domain of tobacco usage detection, showing how tailored approaches can significantly enhance performance in specialized public health monitoring tasks.

\section{Dataset Use Cases}\label{dataset_use_cases}
The PHAD offers a wide range of applications across different fields, primarily focusing on public health, social media analysis, and machine learning research. Below are some key use cases that demonstrate the potential of this dataset.

\subsection{Public Health Research}

\textbf{1. Understanding Tobacco Consumption Trends:} Researchers can analyze the dataset to identify patterns and trends in tobacco consumption, such as the popularity of different tobacco products over time and across various demographics. It can inform public health campaigns and policy-making.

\textbf{2. Evaluating the Impact of Public Health Campaigns:} By monitoring user engagement metrics and sentiment analysis on videos related to public health campaigns, researchers can assess the effectiveness of these campaigns in changing public perceptions and behaviors related to tobacco use.

\textbf{3. Identifying At-Risk Populations:} The dataset can help identify specific demographics that are more likely to engage with tobacco-related content. This information can be used to tailor public health interventions to target these at-risk populations more effectively.

\subsection{Social Media Analysis}

\textbf{1. Sentiment Analysis:} Using the metadata and comments associated with the videos, researchers can perform sentiment analysis to understand public opinion about various tobacco products and regulations. This can help in gauging the public's response to new policies or health warnings.

\textbf{2. Virality and Information Spread:} The dataset can be used to study how tobacco-related content spreads on social media platforms. Researchers can analyze factors that contribute to the virality of such content, including the role of influencers and the impact of social networks.

\textbf{3. Content Moderation and Policy Compliance:} Social media platforms and policymakers can use the dataset to develop and refine content moderation algorithms. This can help in identifying and mitigating the spread of content that violates platform policies or promotes illegal tobacco sales.

\subsection{Machine Learning Research}

\textbf{1. Developing Classification Models:} The dataset provides a rich source of labeled data for training machine learning models to classify different types of tobacco-related content. This can include distinguishing between various tobacco products or identifying promotional versus informational content.

\textbf{2. Enhancing Natural Language Processing (NLP) Techniques:} The textual metadata and comments in the dataset can be used to train and evaluate NLP models. Researchers can develop more sophisticated sentiment analysis, topic modeling, and keyword extraction techniques tailored to public health data.

\textbf{3. Multimodal Learning:} The combination of video, audio, and text data in the dataset offers an opportunity for advancing multimodal learning algorithms. Researchers can explore how integrating different data types can improve the performance of models in tasks such as video classification and behavior analysis.

\subsection{Regulatory and Policy Applications}

\textbf{1. Monitoring Compliance with Advertising Regulations:} Regulators can use the dataset to monitor compliance with advertising regulations related to tobacco products. This can help identify content that may be in violation of laws and take appropriate enforcement actions.

\textbf{2. Informing Policy Decisions:} The insights gained from analyzing the dataset can inform policy decisions regarding tobacco control. For example, understanding which products are most popular among youth can lead to targeted restrictions and educational campaigns.

\textbf{3. Assessing the Effectiveness of Policy Changes:} By analyzing engagement metrics before and after policy changes, researchers can assess the effectiveness of these changes in reducing the prevalence of tobacco-related content and its impact on public health.




\end{appendices}


\bibliography{main}

\end{document}